\documentclass[11pt]{article}
\usepackage[final]{acl}
\pdfoutput=1

\usepackage{latexsym}

\usepackage[T1]{fontenc}

\usepackage[utf8]{inputenc}

\usepackage{microtype}

\usepackage{inconsolata}

\usepackage{graphicx}

\usepackage{times}
\usepackage{latexsym}
\usepackage{placeins}
\usepackage{subcaption}
\usepackage{multirow}
\usepackage[T1]{fontenc}

\usepackage[utf8]{inputenc}

\usepackage{microtype}

\usepackage{inconsolata}
\usepackage{makecell}
\usepackage{booktabs}
\usepackage{array}
\title{An Empirical Study of Recent Large Language Models for Knowledge Graph to Text Generation on Standard Benchmarks}

\pdfoutput=1

\usepackage{times}

\usepackage{placeins}
\usepackage[T1]{fontenc}

\usepackage[utf8]{inputenc}

\usepackage{microtype}

\usepackage{inconsolata}
\usepackage{makecell}

\title{An Empirical Study of Recent Large Language Models for Knowledge Graph to Text Generation on Standard Benchmarks}

\author{
  Priyanka$^{1}$ \quad
  Renaud Delbru$^{2}$ \quad
  Souhail Bakkali$^{3}$ \quad
  Mickaël Coustaty$^{1}$ \quad
  Antoine Doucet$^{1}$ \\[4pt]
  $^{1}$ La Rochelle Université, La Rochelle, France \\
  $^{2}$ Siren, Bordeaux, France \\
  $^{3}$ Université de Rennes, Rennes, France \\[4pt]
  \texttt{\ priyanka.priyanka@univ-lr.fr}
}

\begin{document}
\maketitle
\begin{abstract}
Knowledge graph-to-text (KG-to-text) generation verbalizes a set of knowledge graph triples into fluent natural language. While fine-tuned encoder-decoder models have long been the standard approach, recent LLMs are becoming increasingly competitive. We benchmark recent LLMs against fine-tuned encoder-decoder models, using reference-based metrics and graph-grounded reference-free LLM-based evaluation validated against human annotations. We observed that recent LLMs substantially narrow, and in several settings close, the gap with fine-tuned models on reference-based metrics, while performing better on graph-grounded metrics. LLM-based evaluation scores show limited correlation with reference-based metrics, suggesting that the two approaches capture different aspects of generation quality. Graph-grounded evaluation reveals that prompted models produce fewer contradictions and triple omissions than fine-tuned models, and these errors increase more rapidly with graph size for fine-tuned models, highlighting differences between the two approaches that reference-based metrics may not fully capture.
\end{abstract}

\section{Introduction}
\label{sec:intro}
Knowledge Graphs store information in the form of structured triples as (subject, relation, object). The task of verbalising the knowledge graphs into natural language sentences is a significant and well-studied problem within the natural language generation domain. With the growing capabilities of large language models (LLMs), it is important to understand how effectively recent models perform this task. In this paper advances the evaluation of recent pre-trained models on graph-to-text benchmark datasets, We investigate the performance of five recent model families spanning diverse architectures: dense decoder-only (Gemma-4-E4B-it, Gemma-4-31B-it, \cite{team2026gemma}, Granite-4.2 \cite{granite2026}), mixture-of-experts (Gemma-4-26B-A4B-it \cite{team2026gemma}, Nemotron-3 \cite{blakeman2025nvidia}), encoder-decoder (T5-Gemma2 \cite{zhang2025t5gemma}), closed sourced GPT-5 \cite{singh2025openai} on the following knowledge graph datasets: WebNLG17 \cite{gardent2017webnlg}, WebNLG20 \cite{ferreira20202020} (English subset), and DART \cite{nan2021dart}. \\
For evaluating the performance of KG-to-Text generation systems, the most commonly used metrics are reference-based like BLEU, METEOR, ChrF++, and BLEURT. However, BLEU \cite{papineni2002bleu}, METEOR \cite{banerjee2005meteor}, and ChrF++ \cite{popovic2017chrf++} judge the generated text based on surface similarity to the reference text and do not evaluate its semantic meaningfulness. BLEURT \cite{sellam2020bleurt}, on the other hand, is a learned metric designed to mimic human evaluation. Additionally, we have found that different researchers use different sets and implementations of metrics, which makes it difficult to compare the results with the published State-of-The-Art (SoTA) approaches. Therefore, we have used an elaborate set of metrics consisting of BLEU, METEOR, ChrF++, Rouge-L, BLEURT, BERTScore \cite{zhang2019bertscore}, BARTScore \cite{yuan2021bartscore}, and PARENT \cite{dhingra2019handling} and have categorized them into three different categories based on the function of the metrics. Additionally, we have also studied LLMs as judge category of metrics as source- grounded or reference-free judges and explored how useful and reliable these metrics can be for evaluating LLM-generated text in terms of criteria that cannot be explicitly judged by the standard NLG metrics. We frame the following research questions on the benchmark datasets and try to find the answers:\\
\textbf{RQ1.} Under one scoring pipeline with a multi-reference protocol, where do recent prompted LLMs (GPT-5 mini, Gemma-4-31B, Granite-4.2-30B, Nemotron-3-Ultra) and a LoRA-adapted \cite{hu2021lora} encoder-decoder (T5-Gemma2) model stand against fully fine-tuned T5-large, per metric and per dataset?\\
\textbf{RQ2.} Do the reference-based metric families agree with one another and with a source-grounded judge on these LLM generated outputs?\\
\textbf{RQ3.} What does an LLM judge measure on KG-to-text outputs, when two different LLM judges are compared and validated against human annotation?\\
\textbf{RQ4.} Do fine-tuned and prompted systems make the same kind of errors?\\
Our contributions are: 1) a unified evaluation of five recent model families against fine-tuned T5-large on three benchmarks with eight reference-based metrics, showing that compared with earlier published zero-shot results the gap has narrowed from tens of BLEU points to a few, with parity on neural metrics and a remaining lead for fine-tuning on BLEU, BARTScore and PARENT scores (RQ1); 2) an analysis of metric agreement showing that the reference-based families agree among themselves and not very strongly with a source-grounded judge (RQ2); 3) a two-judge BinEval \cite{cho2026ask} evaluation of all outputs, validated against 220 human-labelled outputs (two model outputs on the same 110 graphs) with adjudicated error types, establishing that the LLM judges reproduce human omission labels triple by triple and detect contradictions and entity corruptions, and that they do not detect unsupported elaboration (RQ3); 4) the finding that, by the human standard, the fine-tuned and the prompted systems have a faithfulness problem in a similar share of outputs but of different kinds: corruption and omission against elaboration(RQ4).
\section{Related work}
\label{sec:Related work}
The early efforts for verbalizing structured data relied on recurrent neural network (RNN) architectures, including Long Short-Term Memory (LSTM) and Gated Recurrent Unit (GRU) models, to encode graph-derived sequences and generate natural language descriptions \cite{song2018graph}; \cite{zhao2020bridging}; \cite{guo2020cyclegt}. Subsequently, Graph Neural Networks (GNN) based models have been employed to use a graph encoder to better capture the structural dependencies and relational information inherent in graph data; \cite{li2021few} and \cite{koncel2019text}. 
More recent approaches, however, have increasingly shifted towards the use of large pre-trained language models, often relying on linearized graph representations as textual input sequences. Recent work has explored both pre-trained sequence-to-sequence architectures \cite{ribeiro2021investigating}; \cite{li2024unifying} and open-source generative language models \cite{he2025evaluating}, \cite{iarosh2025reducing}. The implementations of \citet{ribeiro2021investigating} based on T5 and BART are widely adopted as benchmark baselines for KG-to-Text evaluation.\\
We have used WebNLG17, WebNLG20 and DART as the benchmark datasets. The main reason for choosing these datasets is that they have negligible graph-text alignment noise in their training data i.e. each triple from the graph is described in the corresponding text and there is no information in the text which is not grounded in the corresponding graphs. This property is missing in most of the other KG-to-Text datasets available. For example, GraphNarrative \cite{shi2023hallucination} dataset has on an average 0.62 hallucinated entities and 0.74 hallucinated relations per graph-text pairs, EventNarrative \cite{colas2021eventnarrative} has on an average 0.69 hallucinated entities and 0.70 hallucinated relations, TEKGEN \cite{agarwal2021knowledge} has these numbers as 0.84 hallucinated entities and 0.92 hallucinated relations by construction \cite{he2025evaluating}. Another dataset GenWiki \cite{jin2020genwiki} has 58.1\% unused triplets and 65.1\% unguessable text in the graphs and corresponding text, respectively \cite{kim2026ontology}.\\
While the models finetuned on the respective datasets have consistently achieved higher performance on structured data; \citet{li2024unifying} and \citet{yuan2023evaluating} explored closed-source generative models for generating text from GPT-3 in a zero-shot scenario. Their results showed that the generative models without any fine-tuning fall significantly behind the fine-tuned sequence-to-sequence architectures. Recently, \citet{iarosh2025reducing} evaluated GPT-4o in zero-shot and chain-of-thought (CoT) few-shot settings on WebNLG20 and reported significant performance gains when using detailed prompts. Building on these findings, our work extends the evaluation to a broader set of recent LLM architectures. Evaluation of graph-to-text systems has traditionally relied on reference-based metrics and on human annotation of dimensions such as triple coverage and hallucination. Because human annotation is costly to scale, recent work has turned to LLM-based judges as an automatic alternative \cite{liu2023g}. Rather than asking an LLM for a single holistic rating, CheckEval \cite{lee2025checkeval} and BinEval \cite{cho2026ask} decompose the judgment into a set of binary questions and aggregate the verdicts. A related line of work localises data-text mismatches (omitted, extra and incorrect content), but does so with a fine-tuned verifier rather than LLM prompting \cite{efimov2026xqdt}.  \citet{huidrom2025assessing}, on the other hand, meta-evaluate 29 reference-free metrics against human semantic-consistency ratings on WebNLG. For this study, we have included BinEval and GPTScore \cite{fu2024gptscore} in our evaluation protocol.

\section{Datasets}
\label{sec:Datasets}
 Below, we describe these datasets in detail:\\
\textbf{WebNLG17}. This dataset consists of RDF triples and their corresponding text descriptions released as part of the WebNLG challenge, with the goal of producing a common benchmark to compare “microplanners” for RDF description \cite{gardent2017webnlg}. It has become an important benchmark for KG-to-text generation. The dataset contains graphs having 1 to 7 triples, which are verbalized into a few sentences. All the experiments mentioned in this paper have been performed on the full test subset of the WebNLG17 dataset.\\
\textbf{WebNLG20}.WebNLG dataset has been revised a few times since the first challenge \cite{gardent2017webnlg}. The WebNLG20 \cite{ferreira20202020} dataset has a new category called "company" along with all the categories present in the WebNLG17 version. Consistent with the previous dataset, all the experiments on this dataset have been performed on the full test dataset.\\
\textbf{DART}. DART \cite{nan2021dart} is a large open domain graph to text dataset constructed using various sources: human annotation on
Wikipedia tables from WikiSQL \cite{zhong2017seq2sql} and WikiTableQuestions \cite{pasupat2015compositional}; automatic conversion of questions in WikiSQL into declarative sentences; and inclusion of RDF to text datasets WebNLG 2017 and E2E \cite{duvsek2018findings}. DART thus has triples sourced from diverse domains and serves as a benchmark dataset for graph-to-text generation.

\section{Model Architectures}
\label{models}
The set of model architectures we selected provides checkpoints ranging from 270M-parameter encoder-decoder models to a 550B-parameter sparse model. The architectural difference among these models can be stated as: \textbf{Encoder-decoder vs decoder-only}. We have two models with encoder-decoder architecture, T5 and T5-Gemma2. However, the later is an encoder-decoder model which adapts a pretrained decoder-only Gemma 3 model into an encoder-decoder, and adds tied word embeddings across encoder and decoder and a merged attention module that unifies decoder self- and cross-attention. We use the 270M-270M, 1B-1B and 4B-4B checkpoints. All other models have decoder-only architectures. \textbf{Dense vs Sparse.} Granite-4.2 (with size variants 3B, 8B and 30B) is fully dense decoder-only model. Gemma-4 has size variants 4B-it, 26B-A4B and 31B-it; of which 4B-it and 31B-it are dense decoder-only models whereas 26B-A4B is a mixture-of-experts \cite{shazeer2017outrageously} model with around 26 billion total parameters and 4 billion active parameters with 128 experts and uses interleaved local sliding-window and global attention. The three Nemotron-3 models are sparse with Nemotron-3-Nano-30B-A3B having approximately 3 billion active parameters out of total 30 billion parameters; Nemotron-3-Super-120B-A12B having 12 billion parameters out of 120 billion total parameters and Nemotron-3-Ultra-550B-A55B having 55 billion active parameters out of the total 550 billion parameters and uses interleaved Mamba-2 state-space \cite{dao2024transformers} and MoE layers, with a few self-attention layers.
\textbf{Open vs closed source models}. Out of all the models that we are using, GPT-5's architecture is undisclosed, so it serves as the closed-source reference. We employed three variants of GPT-5: GPT-5-nano, GPT-5-mini, and the full GPT-5 model. 
\section{Experiments}
\label{sec:Experiments}
\subsection{Fine-tuned models}
\label{subsec:Baseline using T5}
We adopted \citet{shi2023hallucination}'s pipeline for fine-tuning T5 for all our T5 experiments. We fine-tuned the different variants of T5 (Small, Base and Large) on the three datasets WebNLG17, WebNLG20 and DART independently, using a learning rate of 3e-5, a batch size of 8 (for training and validation sets), for up to 30 epochs with early stopping (patience of 3) based on validation performance. Source and target sequences are truncated/padded to a maximum length of 384 tokens for both training and generation, with beam search decoding with a beam size of 3 used at inference. The implementation uses linearized knowledge graphs by representing their triples as a sequence of subject–predicate–object elements, concatenated in a fixed order and separated by special delimiter tokens "<H>", "<R>" and "<T>". Training was run on a single A40/A6000 GPU per model. \\We fine-tuned T5Gemma2 (4B-4B variant) for the knowledge-graph-to-text generation task using Low-Rank Adaptation (LoRA). The input format has been kept consistent with the T5 model input setting. We use rank learning rate = 5e-4, r = 32 with scaling factor $\alpha = 64$, dropout = 0.05, applied to the \texttt{q\_proj} and \texttt{v\_proj} modules. This setting was selected via a hyperparameter sweep over $r \in \{8, 16, 32\}$ (with $\alpha = 2r$) and learning rates of $\{1\times10^{-4}, 2\times10^{-4},
3\times10^{-4}, 5\times10^{-4}\}$, averaged over 3 random seeds per configuration to control for initialization variance. Maximum input and target sequence lengths were set according to the each dataset's length distribution.
\subsection{Prompted models }
\label{subsec:GPT-5 experiments}
 For all the decoder-only models and GPT-5, we have used few-shot prompting. In our experiments on these models, following the methodology of \citet{yuan2023evaluating}, each test instance was treated as a separate request, and the model's output was recorded as the generated text.  Instead of using the simple prompt, "generate text from graph," we utilized a more elaborate prompt as shown in Table \ref{tab:[prompt]}. The prompt was inspired by the mistakes noticed when we used the simple prompt, as mentioned in Appendix \ref{appendix:error analysis}, and it is kept the same for all the models.\\We provide the input graphs to the models as a text sequence, we have used the same linearization as used in the case of the T5 and T5-Gemma-2 experiment. Given that most of the open-sourced LLMs we used are reasoning or thinking models, we set the thinking or reasoning level to \textit{"off"} or \textit{"none"} and similarly for GPT-5 we kept reasoning level to \textit{“minimal”}. We also set temperature = 0.0, max tokens = 384 and default top-p value wherever applicable.  We conducted preliminary experiments with GPT-5-mini and Gemma-4-E4B-it on WebNLG test dataset under different few-shot settings to determine the most suitable \(k\)-shot value for the remaining experiments. For selecting few-shot samples, we first use E5-base-V2 \cite{wang2022text} to obtain dense embeddings of the graph linearizations. This model provides an effective balance between retrieval quality and computational efficiency. Each graph was linearized into a flat textual sequence by concatenating the head, relation, and tail tokens of all triples. To mitigate the risk of data leakage, all samples identical to test instances were removed from the training set. For deciding the best sample retrieval strategy, we selected samples for few-shots based on the cosine similarity between the linearized graph embeddings of the test input and the graphs in training set. As \cite{he2025evaluating} mentioned the effectiveness of Difficulty and Diversity of the chosen few shots; we tried Maximal Marginal Relevance \cite{carbonell1998use} as a strategy to balance relevance and non-redundancy in our initial experiments. And as an alternative, we also tried organising both the training and test sets into subsets based on graph triple size. For one-shot retrieval, examples were selected from the subset containing graphs with the same number of triples \(n\) as the test instance. For two-shot retrieval, examples were selected from the \(n\)  and \(n+1\)  triple subsets, while for three-shot retrieval, examples were drawn from the \(n-1\) , \(n\) , and \(n+1\)  triple subsets to introduce structural diversity among the retrieved examples. For test instances belonging to the largest triple-size category, where an \(n+1\)  subset was unavailable, retrieval was restricted to \(n\) triple subset. We discuss the effectiveness of these retrieval strategies in Section \ref{subsec:T5 vs GPT-5}.
\subsection{Evaluation Protocol}
\label{Evaluation}
We evaluate the model outputs from multiple perspectives using a diverse set of evaluation metrics. Based on their underlying methodology, these metrics are categorized into four groups.
\textbf{1) Surface-Level Metrics.}
These metrics evaluate the generated text based on its lexical overlap with the reference text. The metrics included in this category are BLEU \cite{papineni2002bleu}, ChrF++ \cite{popovic2017chrf++}, METEOR \cite{banerjee2005meteor}, and ROUGE-L \cite{lin2004rouge}.We compute corpus-level BLEU and chrF++ scores using the SacreBLEU\cite{post2018call} library. METEOR is computed using the implementation provided in NLTK, while ROUGE-L is computed using the \textit{rouge-score} package.
\textbf{2) Semantic or Neural Metrics.}
These metrics assess the semantic similarity and coherence between the generated text and the reference text, going beyond exact lexical matching. The metrics used in this category include BLEURT \cite{sellam2020bleurt}, BERTScore \cite{zhang2019bertscore}, and BARTScore \cite{yuan2021bartscore}. We compute BERTScore using the \textit{bert-base-uncased} checkpoint, BLEURT using the \textit{BLEURT-20} checkpoint, and BARTScore using the \textit{facebook/bart-large-cnn } checkpoint.
\textbf{3) Alignment-Based Metrics.}
When evaluating Large Language Models for graph-to-text generation, it is essential to determine whether the generated text remains grounded in the source graph. The PARENT \cite{dhingra2019handling} metric addresses this aspect by computing an F1-based score that measures the alignment of the generated text not only with the reference text but also with the source graph.\\
We adopt a multi-reference evaluation protocol for all metrics. For corpus-level metrics such as BLEU and ChrF++, we use SacreBLEU’s native multi-reference evaluation, where each prediction is evaluated against all available references jointly. For sentence-level metrics, including METEOR, ROUGE-L, BLEURT, BARTScore, and PARENT, scores are computed against each reference independently, and the maximum score across references is retained for each instance.\\
\textbf{4) LLM as Judge}
In the task of Graph-to-Text generation, it is very important to judge the outputs explicitly for the faithfulness to the source graph, the coverage of information present in the graph and the fluency of the text as a whole. We leverage LLMs as judges in this task through BinEval \cite{cho2026ask} and GPTScore \cite{fu2024gptscore}. BinEval is a training-free, reference-free framework that decomposes each evaluation criterion into atomic yes/no questions rather than asking an LLM for one holistic score. A meta-prompt first generates the atomic questions from the task prompt, organised by evaluation dimension. An evaluator LLM then checks the output against the source for each question, returning a binary verdict and a short natural-language explanation. The verdicts are aggregated into a score per dimension, so a low score can be traced to the specific failed checks. We evaluated Faithfulness to graph, graph triple Coverage and Fluency of the model outputs using BinEval framework. We use DeepSeekV-4.1-Flash \cite{xu2026deepseek} and GLM-5.3-Flash \cite{glm5team2026glm5vibecodingagentic} with thinking disabled as the judge LLMs. Instead of allowing the BinEval framework to generate binary questions, we provided our own questions for each criterion to the judge LLMs. For each question, the evaluator gives binary output and the final recorded scores are the mean of the total satisfied questions per dimension in the case of Faithfulness and Fluency. The questions provided to the judges are presented in Tables \ref{tab:sf} and \ref{tab:fl}. For Coverage, the final score is the ratio of the sum of the binary score for each triple to the total triples. GPTScore is also a reference-free metric that scores a generated text by the likelihood a pretrained language model assigns to it, conditioned on an instruction. We use GPTScore to get the likelihood of the generated output given the source graph and task instruction. The official implementation of GPTScore is tested with Flan-T5, GPT-2, GPT-3 and OPT family models, however, we have used DeepSeek-V4.1-Flash in non-thinking setting as it is a more recent model and belongs to a different model family compared to the LLMs we are using for text generation. The implementation details for both BinEval and GPTScore are in Appendix \ref{appendix: Bineval and gptscore details}. 

\section{Results}
\label{sec:Results}

\begin{table*}[t!]
\centering
\small
\resizebox{\textwidth}{!}{
\begin{tabular}{lcccccccc}
\hline

\textbf{Model} &
\textbf{BLEU} &
\textbf{METEOR} &
\textbf{ChrF++} &
\textbf{ROUGE-L} &
\textbf{BLEURT} &
\textbf{BERTScore} &
\textbf{BARTScore} &
\textbf{PARENT F1}
\\

\hline

\textbf{T5-large}
& 58.57 {\scriptsize[57.54, 59.56]}
& 0.776 {\scriptsize$\pm 0.007$}
& 72.54 {\scriptsize[71.91, 73.16]}
& 0.735 {\scriptsize$\pm 0.008$}
& 0.820 {\scriptsize$\pm 0.003$}
& 0.883 {\scriptsize$\pm 0.004$}
& -0.962 {\scriptsize$\pm 0.022$}
& 0.693 {\scriptsize$\pm 0.009$}
\\

\textbf{GPT-5 mini}
& 51.94 {\scriptsize[51.01,52.92]}
& 0.761 {\scriptsize$\pm 0.007$}
& 70.70 {\scriptsize[70.13,71.27]}
& 0.702 {\scriptsize$\pm 0.008$}
& 0.814 {\scriptsize$\pm 0.004$}
& 0.866 {\scriptsize$\pm 0.004$}
& -1.117 {\scriptsize$\pm 0.023$}
& 0.649 {\scriptsize$\pm 0.010$}
\\

\textbf{Gemma-4-31B}
& 53.74 {\scriptsize[52.87, 54.62]}
& 0.776 {\scriptsize$\pm 0.007$}
& 71.68 {\scriptsize[71.11, 72.24]}
& 0.717 {\scriptsize$\pm 0.008$}
& 0.820 {\scriptsize$\pm 0.003$}
& 0.873 {\scriptsize$\pm 0.003$}
& -1.035 {\scriptsize$\pm 0.021$}
& 0.667 {\scriptsize$\pm 0.009$}
\\

\textbf{Granite-4.2-30B}
& 52.86 {\scriptsize[51.92, 53.74]}
& 0.770 {\scriptsize$\pm 0.007$}
& 71.49 {\scriptsize[70.95, 72.03]}
& 0.702 {\scriptsize$\pm 0.008$}
& 0.812 {\scriptsize$\pm 0.003$}
& 0.869 {\scriptsize$\pm 0.003$}
& -1.082 {\scriptsize$\pm 0.023$}
& 0.664 {\scriptsize$\pm 0.009$}
\\

\textbf{Nemotron-3-Ultra}
& 55.51 {\scriptsize[54.57,56.42]}
& 0.771 {\scriptsize$\pm 0.007$}
& 71.69 {\scriptsize[71.12,72.21]}
& 0.723 {\scriptsize$\pm 0.008$}
& 0.821 {\scriptsize$\pm 0.003$}
& 0.875 {\scriptsize$\pm 0.003$}
& -1.019 {\scriptsize$\pm 0.021$}
& 0.669 {\scriptsize$\pm 0.009$}
\\

\textbf{T5-Gemma2-4B-4B}
& 55.43 {\scriptsize[54.39, 56.45]}
& 0.765 {\scriptsize$\pm 0.007$}
& 71.36 {\scriptsize[70.74, 71.93]}
& 0.705 {\scriptsize$\pm 0.009$}
& 0.811 {\scriptsize$\pm 0.004$}
& 0.872 {\scriptsize$\pm 0.004$}
& -1.095 {\scriptsize$\pm 0.024$}
& 0.652 {\scriptsize$\pm 0.010$}
\\
\hline

\end{tabular}
}

\caption{Comparison of all models on the WebNLG17 test set. In case of BLEU and ChrF++, we present corpus score and bootstrap 95\% CI. For the other metrics we present the mean scores and 95\% CI through t-distribution}
\label{tab:webnlg-results}

\end{table*}

\begin{table*}[t!]
\centering
\small
\resizebox{\textwidth}{!}{
\begin{tabular}{lcccccccc}
\hline

\textbf{Model} &
\textbf{BLEU} &
\textbf{METEOR} &
\textbf{ChrF++} &
\textbf{ROUGE-L} &
\textbf{BLEURT} &
\textbf{BERTScore} &
\textbf{BARTScore} &
\textbf{PARENT F1}
\\

\hline

\textbf{T5-large}
& 54.23 {\scriptsize[53.30,55.08]}
& 0.745 {\scriptsize$\pm 0.006$}
& 69.48 {\scriptsize[68.93,70.00]}
& 0.692 {\scriptsize$\pm 0.008$}
& 0.802 {\scriptsize$\pm 0.003$}
& 0.863 {\scriptsize$\pm 0.003$}
& -1.079 {\scriptsize$\pm 0.024$}
& 0.649 {\scriptsize$\pm 0.009$}
\\

\textbf{GPT-5-mini}
& 52.08 {\scriptsize[51.20,52.90]}
& 0.745 {\scriptsize$\pm 0.007$}
& 69.19 {\scriptsize[68.66,69.73]}
& 0.683 {\scriptsize$\pm 0.008$}
& 0.807 {\scriptsize$\pm 0.003$}
& 0.859 {\scriptsize$\pm 0.003$}
& -1.121 {\scriptsize$\pm 0.021$}
& 0.637 {\scriptsize$\pm 0.009$}
\\

\textbf{Gemma-4-31B}
& 54.85 {\scriptsize[53.95, 55.69]}
& 0.764 {\scriptsize$\pm 0.007$}
& 70.65 {\scriptsize[70.10, 71.21]}
& 0.701 {\scriptsize$\pm 0.008$}
& 0.817 {\scriptsize$\pm 0.003$}
& 0.869 {\scriptsize$\pm 0.003$}
& -1.045 {\scriptsize$\pm 0.021$}
& 0.658 {\scriptsize$\pm 0.009$}
\\

\textbf{Granite-4.2-30B }
& 51.89 {\scriptsize[51.06,52.77]}
& 0.749 {\scriptsize$\pm 0.006$}
& 69.55 {\scriptsize[69.04,70.06]}
& 0.675 {\scriptsize$\pm 0.008$}
& 0.805 {\scriptsize$\pm 0.003$}
& 0.857 {\scriptsize$\pm 0.003$}
& -1.136 {\scriptsize$\pm 0.023$}
& 0.640 {\scriptsize$\pm 0.009$}
\\

\textbf{Nemotron-3-Ultra}
& 53.99 {\scriptsize[53.10,54.87]}
& 0.749 {\scriptsize$\pm 0.007$}
& 69.73 {\scriptsize[69.18,70.28]}
& 0.691 {\scriptsize$\pm 0.008$}
& 0.812 {\scriptsize$\pm 0.003$}
& 0.863 {\scriptsize$\pm 0.003$}
& -1.035 {\scriptsize$\pm 0.020$}
& 0.638 {\scriptsize$\pm 0.009$}
\\

\textbf{T5-Gemma2-4B-4B}
& 51.45 {\scriptsize[50.52,52.37]}
& 0.728 {\scriptsize$\pm 0.007$}
& 67.93 {\scriptsize[67.37,68.48]}
& 0.672 {\scriptsize$\pm 0.008$}
& 0.793 {\scriptsize$\pm 0.004$}
& 0.852 {\scriptsize$\pm 0.004$}
& -1.167{\scriptsize$\pm 0.025$}
& 0.633 {\scriptsize$\pm 0.009$}
\\
\hline

\end{tabular}
}

\caption{Comparison of all models on the WebNLG20 test set. In case of BLEU and ChrF++, we present corpus score and bootstrap 95\% CI. For the other metrics we present the mean scores and 95\% CI through t-distribution}
\label{tab:webnlg20-results}

\end{table*}
\begin{table*}[t!]
\centering
\small
\resizebox{\textwidth}{!}{
\begin{tabular}{lcccccccc}
\hline

\textbf{Model} &
\textbf{BLEU} &
\textbf{METEOR} &
\textbf{ChrF++} &
\textbf{ROUGE-L} &
\textbf{BLEURT} &
\textbf{BERTScore} &
\textbf{BARTScore} &
\textbf{PARENT F1}
\\

\hline

\textbf{T5-large}
& 47.87 {\scriptsize [47.21, 48.51]}
& 0.706 {\scriptsize$\pm 0.005$}
& 65.11 {\scriptsize[64.69,65.53]}
& 0.647 {\scriptsize$\pm 0.005$}
& 0.779 {\scriptsize$\pm 0.003$}
& 0.844 {\scriptsize$\pm 0.003$}
& -1.182 {\scriptsize$\pm 0.017$}
& 0.609 {\scriptsize$\pm 0.006$}
\\

\textbf{GPT-5-mini}
& 41.80 {\scriptsize[41.22,42.35]}
& 0.686 {\scriptsize$\pm 0.005$}
& 63.40 {\scriptsize[63.02,63.80]}
& 0.615 {\scriptsize$\pm 0.005$}
& 0.773 {\scriptsize$\pm 0.003$}
& 0.824 {\scriptsize$\pm 0.002$}
& -1.427 {\scriptsize$\pm 0.017$}
& 0.588 {\scriptsize$\pm 0.006$}
\\

\textbf{Gemma-4-31B}
& 43.95 {\scriptsize[43.35,44.50]}
& 0.696 {\scriptsize$\pm 0.005$}
& 64.21 {\scriptsize[63.83,64.60]}
& 0.632 {\scriptsize$\pm 0.005$}
& 0.784 {\scriptsize$\pm 0.002$}
& 0.836 {\scriptsize$\pm 0.002$}
& -1.279 {\scriptsize$\pm 0.016$}
& 0.609 {\scriptsize$\pm 0.006$}
\\

\textbf{Granite-4.2-30B}
& 42.81 {\scriptsize[42.20,43.43]}
& 0.691 {\scriptsize$\pm 0.005$}
& 63.93 {\scriptsize[63.54,64.34]}
& 0.617 {\scriptsize$\pm 0.005$}
& 0.777 {\scriptsize$\pm 0.003$}
& 0.831 {\scriptsize$\pm 0.003$}
& -1.381 {\scriptsize$\pm 0.017$}
& 0.597 {\scriptsize$\pm 0.006$}
\\

\textbf{Nemotron-3-Ultra}
& 45.04 {\scriptsize[44.40,45.69]}
& 0.689 {\scriptsize$\pm 0.005$}
& 62.82 {\scriptsize[62.40,63.24]}
& 0.635 {\scriptsize$\pm 0.005$}
& 0.783 {\scriptsize$\pm 0.003$}
& 0.836 {\scriptsize$\pm 0.003$}
& -1.262 {\scriptsize$\pm 0.016$}
& 0.609 {\scriptsize$\pm 0.006$}
\\

\textbf{T5-Gemma2-4B-4B}
& 45.04 {\scriptsize[44.40,45.69]}
& 0.685 {\scriptsize$\pm 0.005$}
& 62.83 {\scriptsize[62.40,63.24]}
& 0.623 {\scriptsize$\pm 0.006$}
& 0.762 {\scriptsize$\pm 0.003$}
& 0.833 {\scriptsize$\pm 0.003$}
& -1.306 {\scriptsize$\pm 0.018$}
& 0.598 {\scriptsize$\pm 0.006$}
\\
\hline

\end{tabular}
}

\caption{Comparison of all models on the DART test set. In case of BLEU and ChrF++, we present corpus score and bootstrap 95\% CI. For the other metrics we present the mean scores and 95\% CI through t-distribution}
\label{tab:dart-results}

\end{table*}

\subsection{Standing of recent models on reference based evaluation (RQ1)}
\label{subsec:T5 vs GPT-5}
\begin{figure*}[t]
    \centering
    \includegraphics[width=0.75\textwidth,height=0.7\textheight,keepaspectratio]{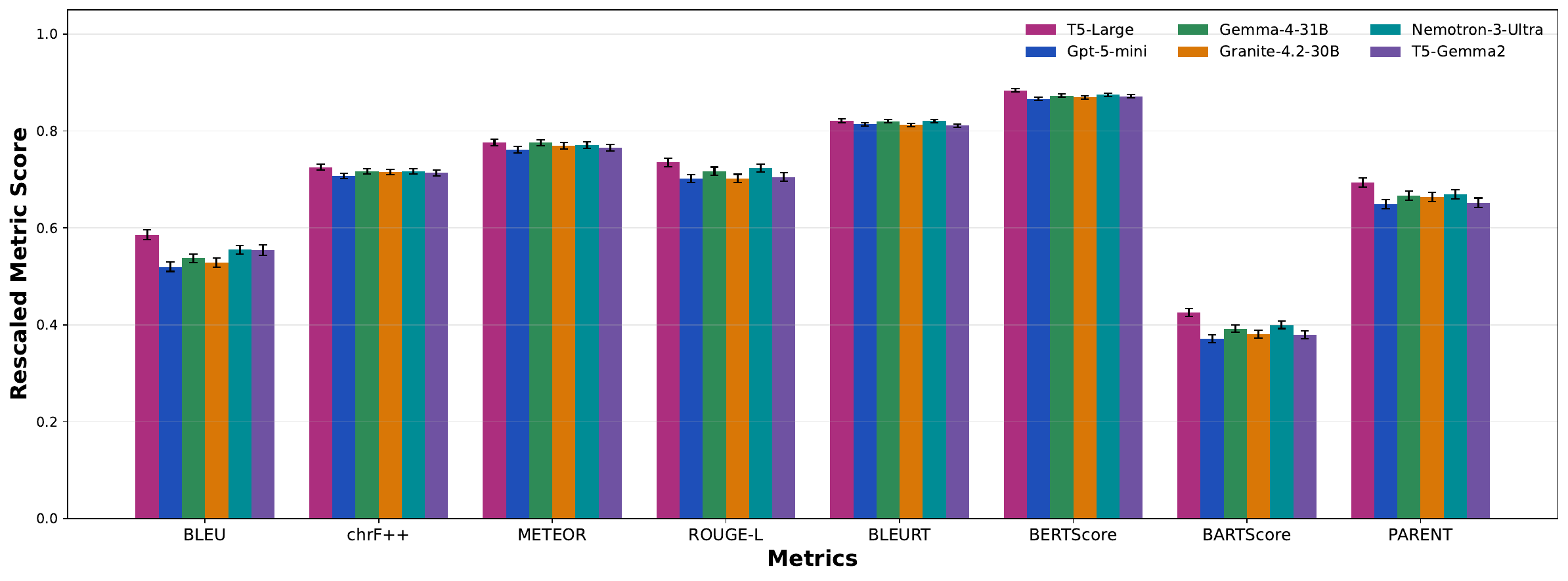}
    \caption{Mean metric scores per model on WebNLG17, with 95\% confidence intervals. For visualization, metric scores are rescaled to \([0,1]\): BARTScore is transformed via $\exp(\cdot)$; BLEU and chrF++ are divided by 100 to match the \([0,1]\) range of the remaining metrics, which require no rescaling.}
    \label{fig:webnlg}
\end{figure*}

\begin{figure*}[t]
    \centering
    \includegraphics[width=0.75\textwidth,height=0.7\textheight,keepaspectratio]{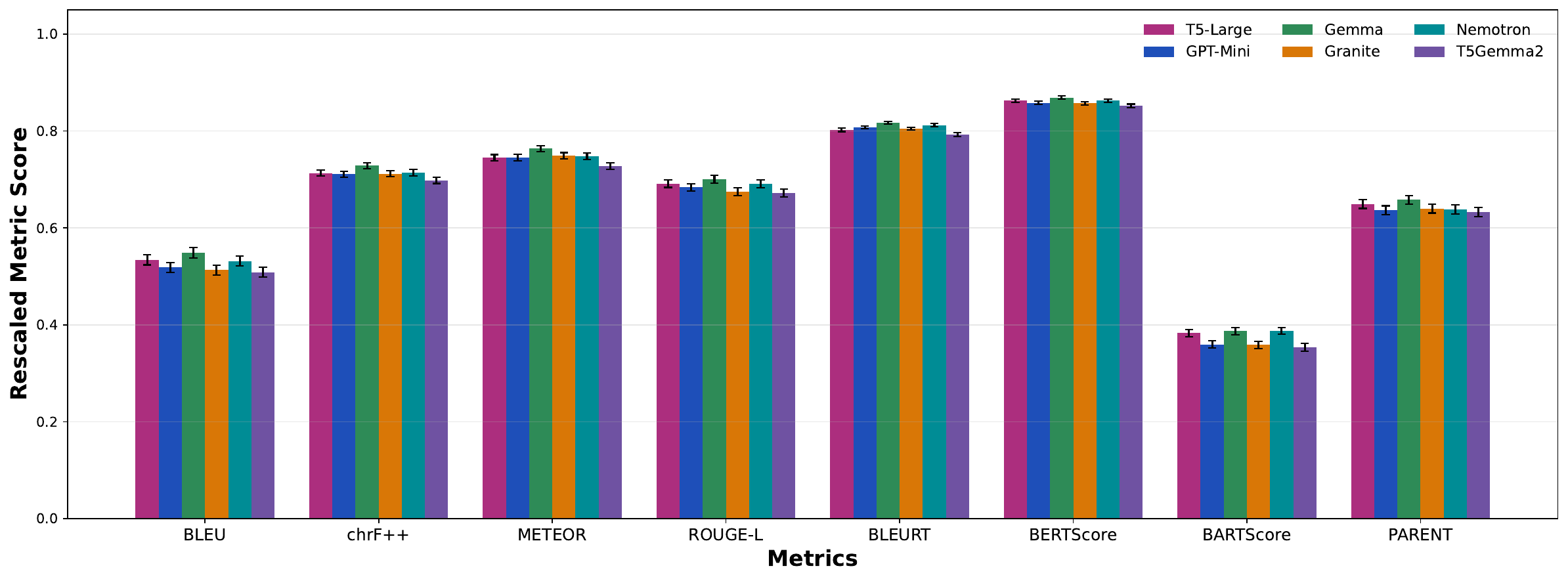}
    \caption{Mean metric scores per model on WebNLG20, with 95\% confidence intervals. For visualization, metric scores are rescaled to \([0,1]\): BARTScore is transformed via $\exp(\cdot)$; BLEU and chrF++ are divided by 100 to match the \([0,1]\) range of the remaining metrics, which require no rescaling.}
    \label{fig:webnlg+}
\end{figure*}

\begin{figure*}[t]
    \centering
    \includegraphics[width=0.75\textwidth,height=0.7\textheight,keepaspectratio]{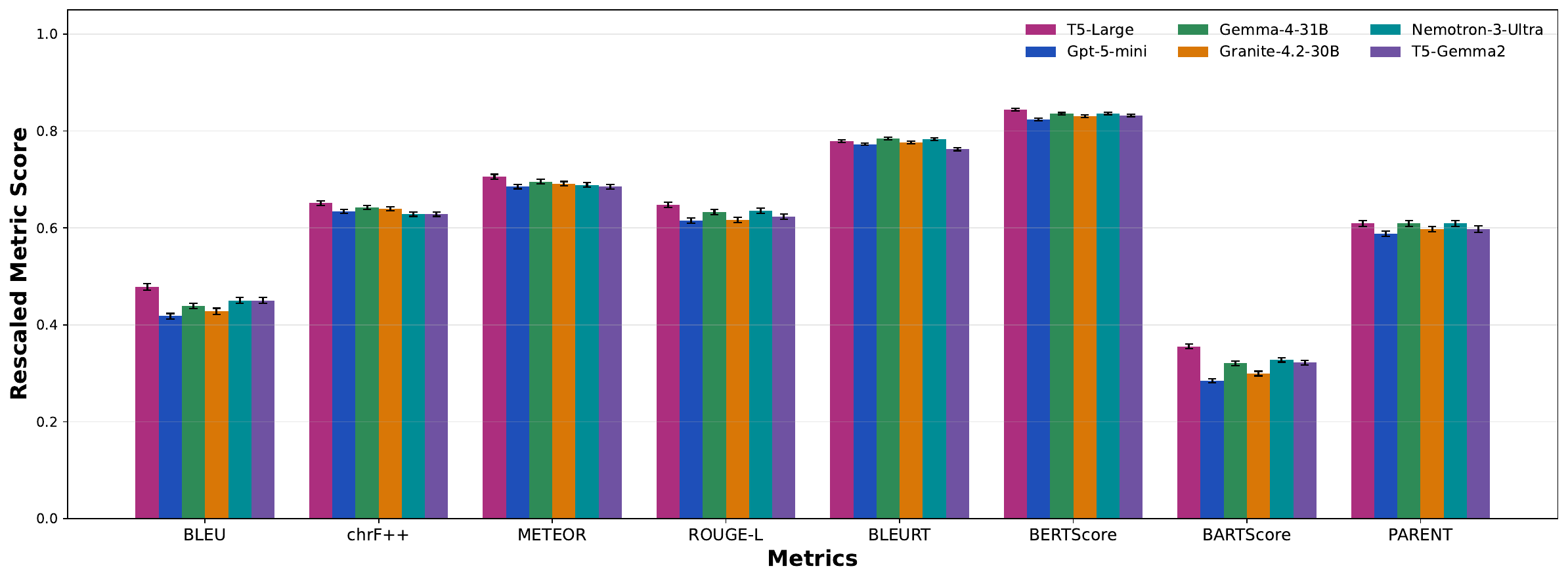}
    \caption{Mean metric scores per model on DART, with 95\% confidence intervals. For visualization, metric scores are rescaled to \([0,1]\): BARTScore is transformed via $\exp(\cdot)$; BLEU and chrF++ are divided by 100 to match the \([0,1]\) range of the remaining metrics, which require no rescaling.}
    \label{fig:dart}
\end{figure*}

\begin{figure}[t]
    \centering
    \includegraphics[width=\linewidth]{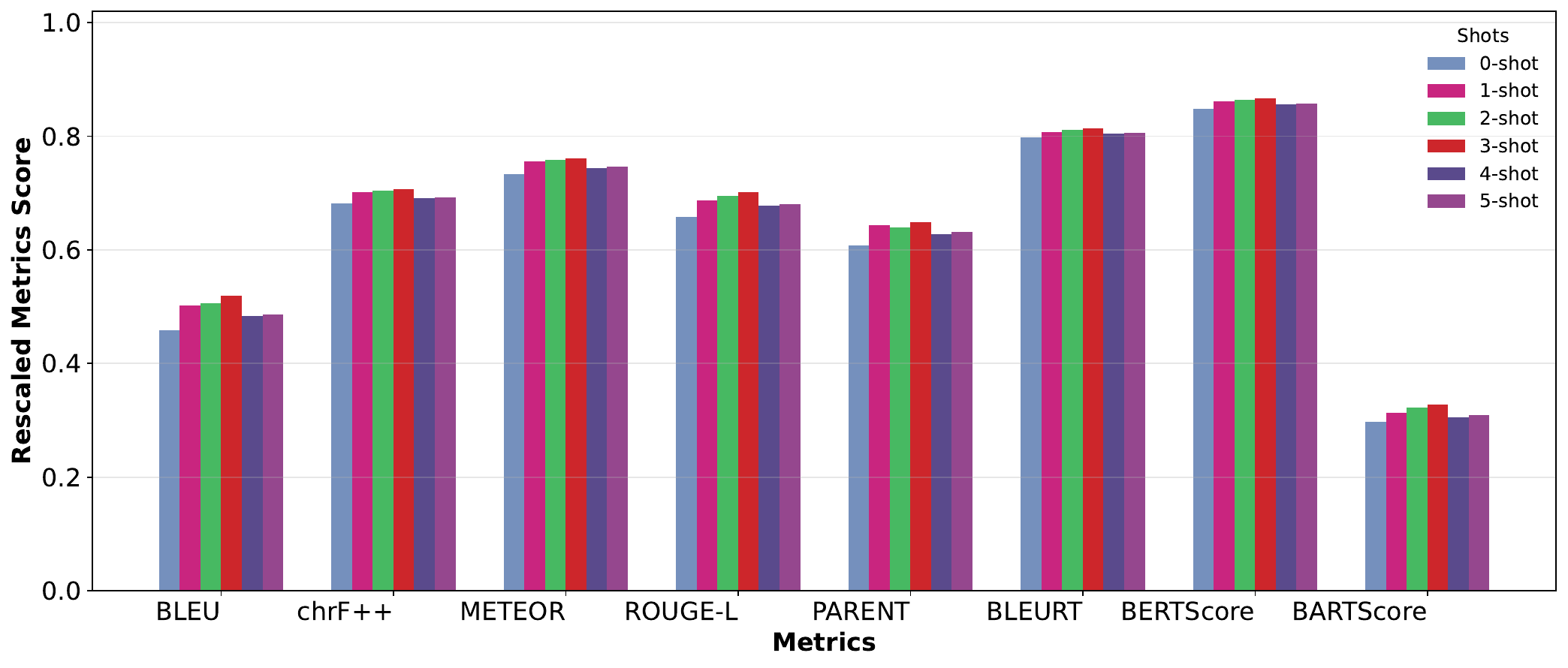}
    \caption{Few shot result comparison on GPT-5-mini on WebNLG17. For visualization, metric scores are rescaled to \([0,1]\): BARTScore is transformed via $\exp(\cdot)$; BLEU and chrF++ are divided by 100 to match the \([0,1]\) range of the remaining metrics, which require no rescaling. }
    \label{fig:gptfewshots}
\end{figure}

\begin{figure}[t]
    \centering
    \includegraphics[width=\linewidth]{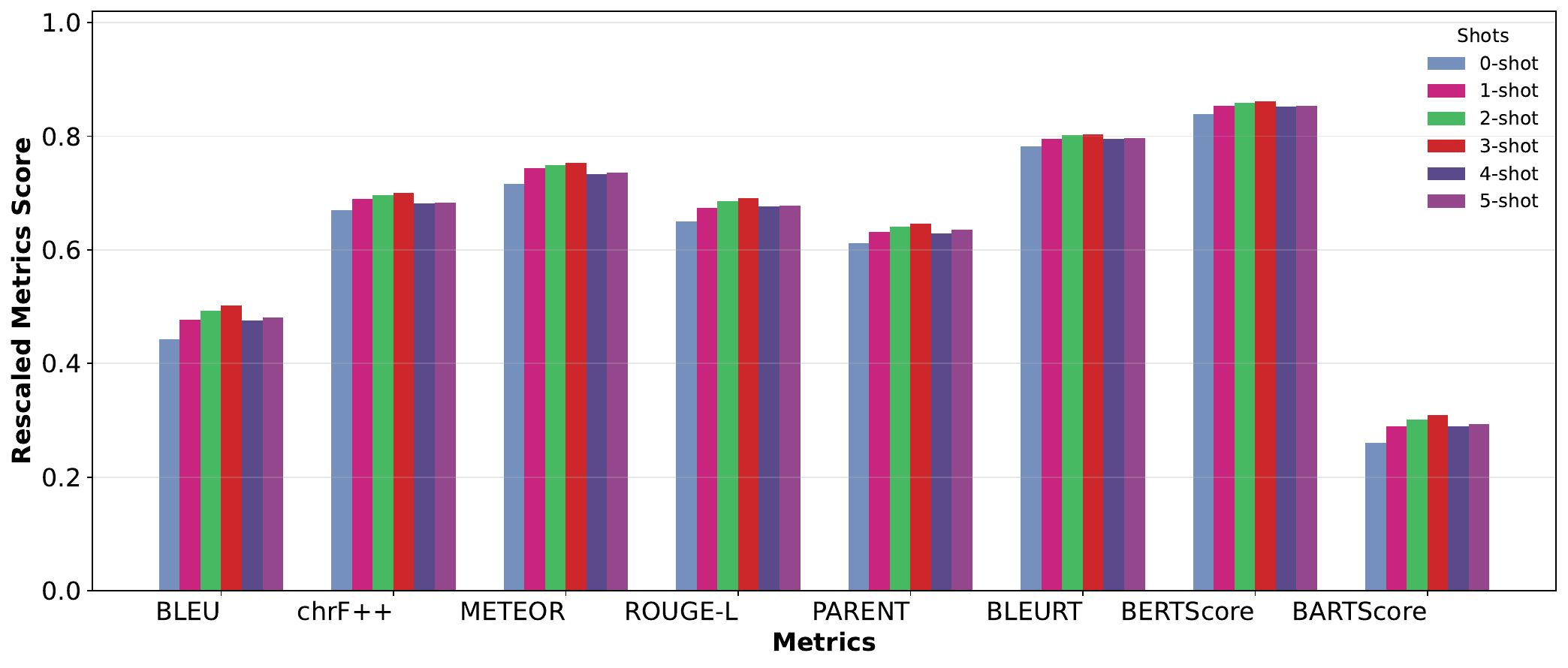}
    \caption{Few shot result comparison on Gemma-4-E4B-it on WebNLG17. For visualization, metric scores are rescaled to \([0,1]\): BARTScore is transformed via $\exp(\cdot)$; BLEU and chrF++ are divided by 100 to match the \([0,1]\) range of the remaining metrics, which require no rescaling.}
    \label{fig:gemmafewshots}
\end{figure}

The results obtained on all the models are presented in Tables \ref{tab:webnlg-results}, \ref{tab:webnlg20-results} and \ref{tab:dart-results}. 95\% Confidence Intervals (CIs) are computed via paired bootstrap resampling (\cite{koehn2004statistical}, N=1,000) for BLEU and ChrF++, and via Student’s t over per-sample scores for all other metrics.\\
For all the models, we observed an increase in performance when moving from smaller variants to larger variants within the same family so the final scores presented in this study are on the largest variants of each model family (T5-large, Gemma-4-31B, Granite-4.2-30B, Nemotron-3-Ultra-550B-55B and T5-Gemma2-4B-4B). However, the performance gain achieved by GPT-5 full was not found to be very significant compared to GPT-5-mini. Considering that GPT-5 full costs more than GPT-5-mini and nano, we have conducted the reported GPT-5 experiments on GPT-5-mini. In section \ref{subsec:GPT-5 experiments}, we described the different ways of sample retrieval for few-shot selection.  The approach of using Maximal Marginal Relevance for few-shot retrieval showed some performance gain against cosine similarity-based retrieval with \(\lambda=0.7\) but the gain was very small. We also experimented with sample retrieval based on graph triple size, but the performance of GPT-5-mini using this retrieval strategy did not surpass the simpler cosine similarity-based retrieval approach. Consequently, we adopted cosine similarity-based retrieval for all few-shot experiments reported in this work.\\
To select the number of shots for few-shot experiments, we evaluated GPT-5-mini and Gemma-4-E4B-it on WebNLG17 test dataset in \(k-shot\) scenarios with \(k\) ranging from 0 to 5. The increase in performance versus \(k\) on GPT-5-mini and Gemma-4-E4B-it is shown in Figures \ref{fig:gptfewshots} and \ref{fig:gemmafewshots} respectively. The findings are consistent with those of \citet{he2025evaluating}. There is a performance increase when we go from \(k=0\) to \(k=3\). However, after \(k=3\), there is a decrease in performance. The reported results in this paper come from \(3\)-shot prompting for all three datasets on all prompted LLMs.\\
Coming to model comparison, on WebNLG17, T5-large, 
has remained the best-performing model among almost all of the metrics where Nemotron-3-Ultra scored equal METEOR and BLEURT,  and lagging by 1 to 2 points on the rest of the metrics (see Table \ref{tab:webnlg-results} and Figure \ref{fig:webnlg}); however, for WebNLG20 dataset, Gemma-4-31B-it has performed better than T5-large for all the metrics and Nemotron-3-Ultra has scored the best BARTScore while being within 1 point range of Gemma-4 and T5-large (Table \ref{tab:webnlg20-results} and Figure \ref{fig:webnlg+}). On DART, T5-large remained the best model on all metrics except on BLEURT where Gemma-4-31B-it and Nemotron-3-Ultra reported the best scores (Table \ref{tab:dart-results} and Figure \ref{fig:dart}).

\begin{table*}[t]
\centering
\small
\setlength{\tabcolsep}{4pt}
\resizebox{\textwidth}{!}{%
\begin{tabular}{lcccccccccccc}
\toprule
& \multicolumn{6}{c}{Faithfulness}
& \multicolumn{6}{c}{Coverage} \\
\cmidrule(lr){2-7}
\cmidrule(lr){8-13}
System
& \multicolumn{2}{c}{WebNLG17}
& \multicolumn{2}{c}{WebNLG20}
& \multicolumn{2}{c}{DART}
& \multicolumn{2}{c}{WebNLG17}
& \multicolumn{2}{c}{WebNLG20}
& \multicolumn{2}{c}{DART} \\
\cmidrule(lr){2-3}
\cmidrule(lr){4-5}
\cmidrule(lr){6-7}
\cmidrule(lr){8-9}
\cmidrule(lr){10-11}
\cmidrule(lr){12-13}
& DeepSeek & GLM
& DeepSeek & GLM
& DeepSeek & GLM
& DeepSeek & GLM
& DeepSeek & GLM
& DeepSeek & GLM \\
\midrule
T5-Large        & 0.878 & 0.925 & 0.882 &0.921 & 0.860 & 0.918 & 0.956 & 0.965 & 0.948 & 0.956 & 0.951 & 0.963 \\
T5-Gemma 2     & 0.840 & 0.890 & 0.824 & 0.868 & 0.819 & 0.853 & 0.954 & 0.963 & 0.936 & 0.945 & 0.912 & 0.914 \\
GPT-5-mini        & 0.956 & 0.983 & 0.959 & 0.989 & 0.971 & 0.986 & 0.992 & 0.997 & 0.993 & 0.996 & 0.993 & 0.997 \\
Granite-4.2     & 0.943 & 0.969 & 0.946 & 0.968 & 0.949 & 0.977 & 0.989 & 0.993 & 0.989 & 0.992 & 0.987 & 0.994 \\
Nemotron-3-Ultra & 0.952 & 0.983 & 0.952 & 0.989 & 0.955 & 0.982 & 0.991 & 0.996 & 0.988 & 0.996 & 0.991 & 0.994 \\
Gemma-4    & 0.974 & 0.988 & 0.970 & 0.993 & 0.980 & 0.989 & 0.992 & 0.996 & 0.992 & 0.998 & 0.993 & 0.997 \\
\bottomrule
\end{tabular}%
}
\caption{Faithfulness and coverage scores across six systems, three datasets, and two LLM judges. DeepSeek stands for DeepSeek-V-4.1-Flash and GLM stands for GLM-5.3-Flash.}
\label{tab:faithfulness_coverage}
\end{table*}

\begin{table*}[t]
\centering
\small
\setlength{\tabcolsep}{6pt}
\renewcommand{\arraystretch}{1.1}
\resizebox{\textwidth}{!}{%
\begin{tabular}{l*{10}{c}}

\toprule
& \multicolumn{2}{c}{\makecell{Q1: Unsupported\\facts (\%)}}
& \multicolumn{2}{c}{\makecell{Q2: Incorrect Names\\and Values (\%)}}
& \multicolumn{2}{c}{\makecell {Q3: Incorrect\\Relations (\%)}}
& \multicolumn{2}{c}{Any question (\%)}
& \multicolumn{2}{c}{All three} \\
\cmidrule(lr){2-3}
\cmidrule(lr){4-5}
\cmidrule(lr){6-7}
\cmidrule(lr){8-9}
\cmidrule(lr){10-11}
System
& DeepSeek & GLM
& DeepSeek & GLM
& DeepSeek & GLM
& DeepSeek & GLM
& DeepSeek & GLM \\
\midrule
T5-Large          & 12.5 & 7.8 & 13.4 & 7.8 & 10.5 & 6.8 & 17.5 & 12.1 & 146 & 72 \\
T5-Gemma2    & 15.4 & 11.2 & 19.4 & 12.9 & 13.0 & 8.8 & 23.2 & 16.4 & 198 & 119 \\
Gemma-4-31B       & 3.0 & 1.5 & 2.7 & 1.0 & 2.3 & 1.2 & 4.5 & 2.4 & 23 & 7 \\
Granite-4.2-30B   & 6.3 & 4.2 & 6.5 & 2.5 & 4.4 & 2.5 & 9.7 & 5.3 & 46 & 24 \\
GPT-5 mini        & 3.7 & 2.2 & 6.6 & 1.8 & 2.9 & 1.1 & 9.2 & 3.4 & 20 & 7 \\
Nemotron-3-Ultra  & 4.9 & 2.3 & 5.8 & 1.7 & 3.6 & 1.2 & 8.5 & 3.5 & 37 & 10 \\
\bottomrule
\end{tabular}%
}
\caption{Share of outputs per model (system) where each judge answered "no" to the corresponding question.}
\label{per question faithful}
\vspace{3pt}
\begin{minipage}{0.95\textwidth}

\end{minipage}
\end{table*}

\begin{table*}[t!]
\centering
\footnotesize
\setlength{\tabcolsep}{3pt}
\renewcommand{\arraystretch}{0.95}

\begin{tabular}{lcccc}
\hline
\textbf{Dimension} &
\textbf{System} &
\textbf{Human Negatives} &
\textbf{\makecell{DeepSeek\\Precision, Recall, $\kappa$}} &
\textbf{\makecell{GLM\\Precision, Recall, $\kappa$}} \\
\hline

\textbf{Faithfulness}
& T5-Large
& 14
& 11/21, 11/14, 0.56
& 11/14, 11/14, 0.75
\\

\textbf{Faithfulness}
& Nemotron-3-Ultra
& 9
& 1/6, 1/9, 0.07
& 0/2, 0/9, -0.03
\\

\textbf{Coverage}
& T5-Large
& 11
& 10/16, 10/11
& 9/13, 9/11
\\
\textbf{Coverage}
& Nemotron-3-Ultra
& 1
& 1/2, 1/1
& 1/1, 1/1
\\
\hline
\end{tabular}

\caption{LLM judge agreement with manual annotation on 110 WebNLG T5-Large and Nemotron-3-Ultra outputs ($\kappa$ denotes Cohen's kappa).}
\label{tab:manual annotation}

\end{table*}

\subsection{LLMs-as-judge results (RQ3)}
\label{subsec:LLM-as-judge}
The evaluation results from LLMs as judge for all the three datasets on Faithfulness and Coverage is presented in Table \ref{tab:faithfulness_coverage} and for WebNLG17 and WebNLG20 on Fluency in Table \ref{tab:fluency}. Both the evaluator LLMs have ranked the models similarly, with the prompted models being better than the fine-tuned models, which is not consistent with the ranking of models given by reference-based metrics, where the encoder-decoder model T5 is ranked the best in most scenarios. Another observation is that GLM-5.3-Flash as an evaluator, is more lenient than DeepSeek-V4.1-Flash on Faithfulness and Coverage and stricter on Fluency, so rankings and differences are comparable across judges, and absolute scores are not; Fluency is saturated and differs more between judges than between systems, so no conclusion is drawn from it in this paper. \\
On careful study of per-question failures on the full WebNLG17 outputs on all the generator models, we observed that fine-tuned models (T5 and LoRA-tuned T5-Gemma2) separate cleanly from the prompted systems under both the LLM judges. The two fine-tuned systems have a faithfulness failure on 12\% to 23\% of their outputs, and the four prompted systems on 2\% to 10\% (Table \ref{per question faithful}). A T5 or T5-Gemma2 output failure is usually a whole-output corruption, whereas a prompted model output usually fails on a single question. Among failed outputs, the share of all-question failure is 32\% to 46\% for the two fine-tuned systems and 11\% to 27\% for the four prompted ones. In counts: 146 and 198 outputs under DeepSeek for T5-large and T5-Gemma2, against 20 to 46 for the prompted systems. The two judges agree with each other more on the fine-tuned systems (Cohen’s kappa 0.62 and 0.68 over all binary verdicts for each model’s outputs) compared to the prompted ones (0.32 to 0.45). Where real errors exist, the judges converge; where they do not, each judge raises its own spurious flags. Also, the errors grow with graph size, and much faster for the fine-tuned models. From graphs of one or two triples to graphs of four or more, the share of outputs with a faithfulness failure goes from 8\% to 26\% for T5-large and from 9\% to 38\% for T5-Gemma2 under DeepSeek, and the share with a missing triple from 3\% to 27\% for both (Tables \ref{tab:faithfulness_by_triples} and \ref{tab:coverage_by_triples}. 
GPTScore on the other hand has remained largely inconclusive as clear from the Table \ref{tab:gptscore}.
\subsection{Validation against human annotations (RQ3, RQ4)}
\label{subsec:human annotations}
To establish what the LLM judge scores measure, one author annotated the outputs of T5-large and Nemotron-3-Ultra on the same 110 WebNLG17 graphs (with graphs containing triples 1 to 7, with at least five graphs of sizes 6 and 7), answering the three Faithfulness questions provided in Table \ref{tab:sf} for each output and recording omitted triples. A second author then analysed every failed question and every LLM judge verdict that the annotator had passed, with the judges’ explanations visible; the decisions of the second author are final.\\
According to the annotator, the failed questions fall into three kinds: Unsupported Elaborations (where the text adds plausible content that the graph does not state, failure of question 1 from Table \ref{tab:sf}), Entity Corruptions (misspelt or altered entity names and values, failure of question 2), and Contradictions (wrong or reversed relations, failure of question 3). Contradictions and entity corruptions occur only in T5-large outputs on the annotated sample (10 out of 110 outputs); DeepSeek detects all ten such failed questions and GLM eight. Unsupported Elaborations are the only failure cases of Nemotron-3-Ultra (9 of 110 outputs) and also occur in T5-large (3 out of 110 outputs) on the annotated sample; the judges flagged only one of these twelve while on the full dataset, the judges do catch some unsupported elaborations of one specific kind: a relation reworded with more meaning than the graph gives (for example, the output mentioning “plays for” for the relation "club", “county seat” for relation "largest city", and “official language” for relation "language"; etc.). At the output level, agreement with the adjudicated labels on T5-large is Cohen’s kappa 0.56 for DeepSeek (precision 11/21, recall 11/14) and 0.75 for GLM (11/14, 11/14) (Table \ref{tab:manual annotation}).\\
For graph triple coverage, the judges reproduce the annotator’s omissions at the level of the individual triple. On T5-large, which omits at least one triple in 11 of 110 outputs (mostly on graphs with four or more triples), DeepSeek recovers 10 and GLM 9, naming the same missing triples; their remaining flags are triples that were rendered incorrectly or expressed without a qualifier. Nemotron-3-Ultra omits a triple once, and both judges find it. The coverage scores in the judge figures can be read as triple omission rates.\\
By the annotator’s standard the two systems have flagged Faithfulness failure in a similar share of outputs (14 and 9 of 110, paired sign test p = 0.36) but of entirely different kinds. The LLM judge Faithfulness scores should therefore be read as rates of hard faithfulness errors (contradiction and entity corruption), on which the prompted models are clearly better. Table \ref{tab:manual annotation} summarises the validation.
\subsection{Agreement between metrics (RQ2)}
\label{subsec:Agreement between metrics}
\begin{figure*}[t]
    \centering

    \begin{subfigure}[t]{0.48\textwidth}
        \centering
        \includegraphics[width=\linewidth]{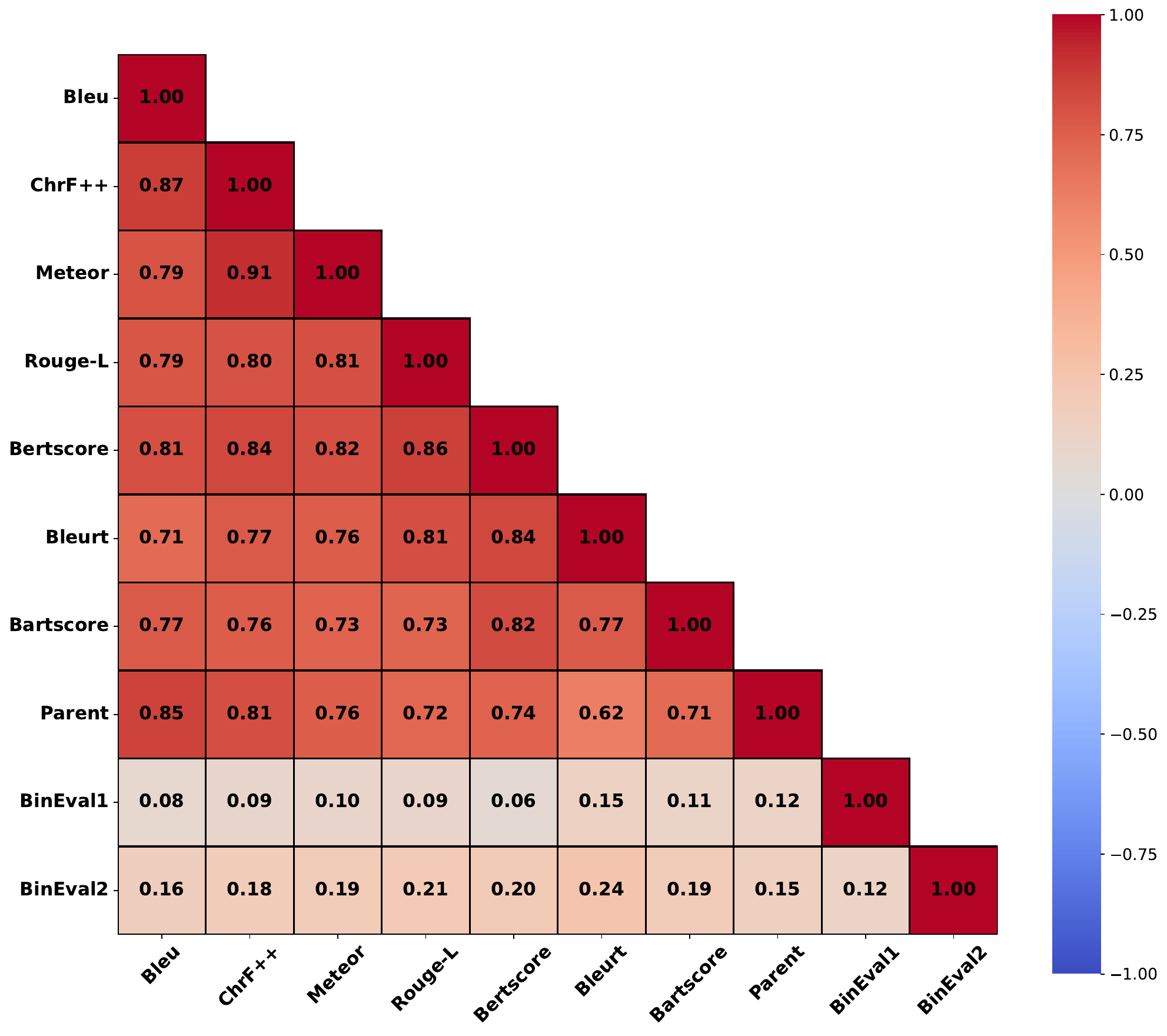}
        \caption{GPT-5-mini}
        \label{fig:gpt_correlation}
    \end{subfigure}
    \hfill
    \begin{subfigure}[t]{0.48\textwidth}
        \centering
        \includegraphics[width=\linewidth]{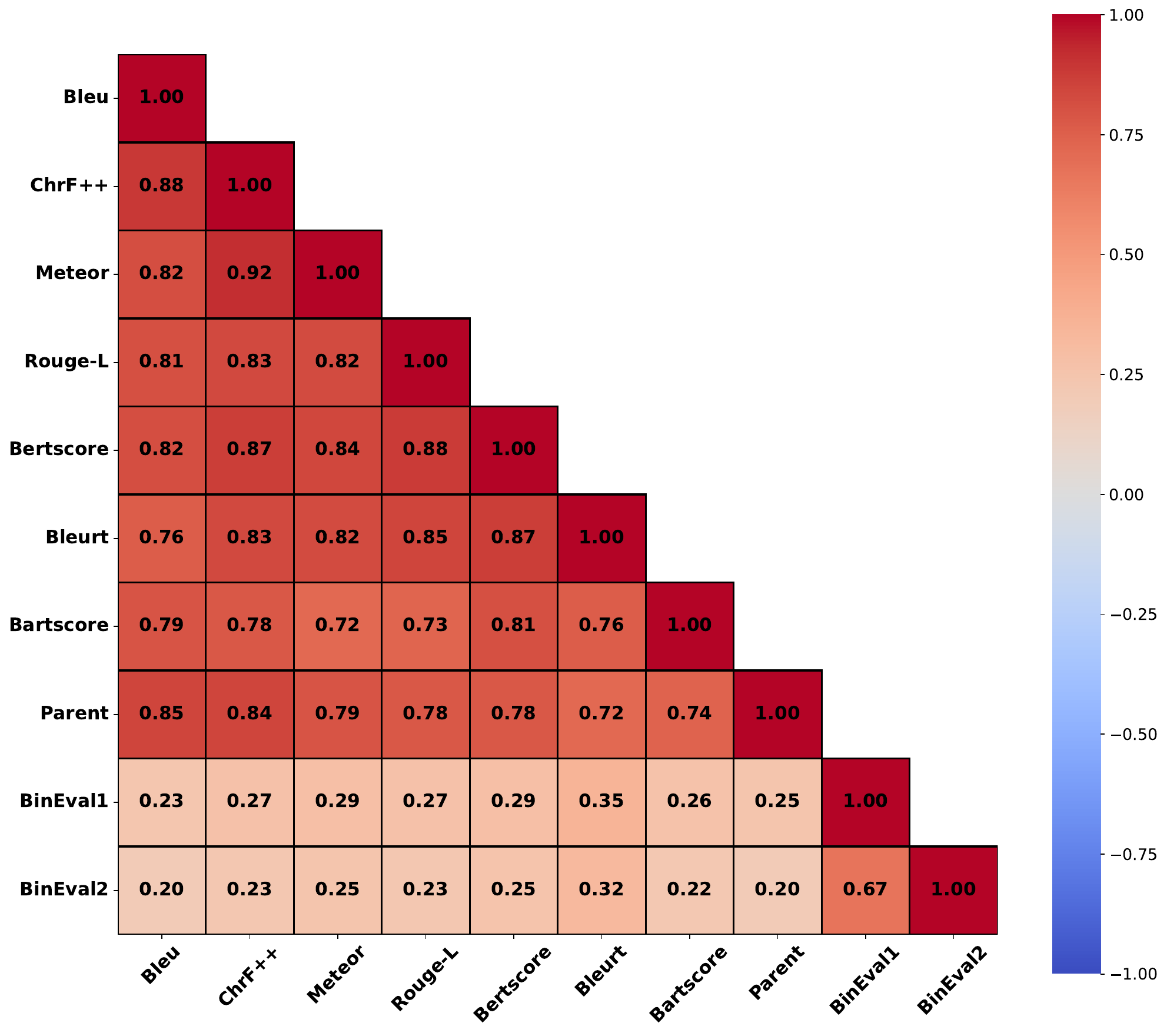}
        \caption{T5-Large}
        \label{fig:t5_correlation}
    \end{subfigure}

    \vspace{0.5em}

    \begin{subfigure}[t]{0.48\textwidth}
        \centering
        \includegraphics[width=\linewidth]{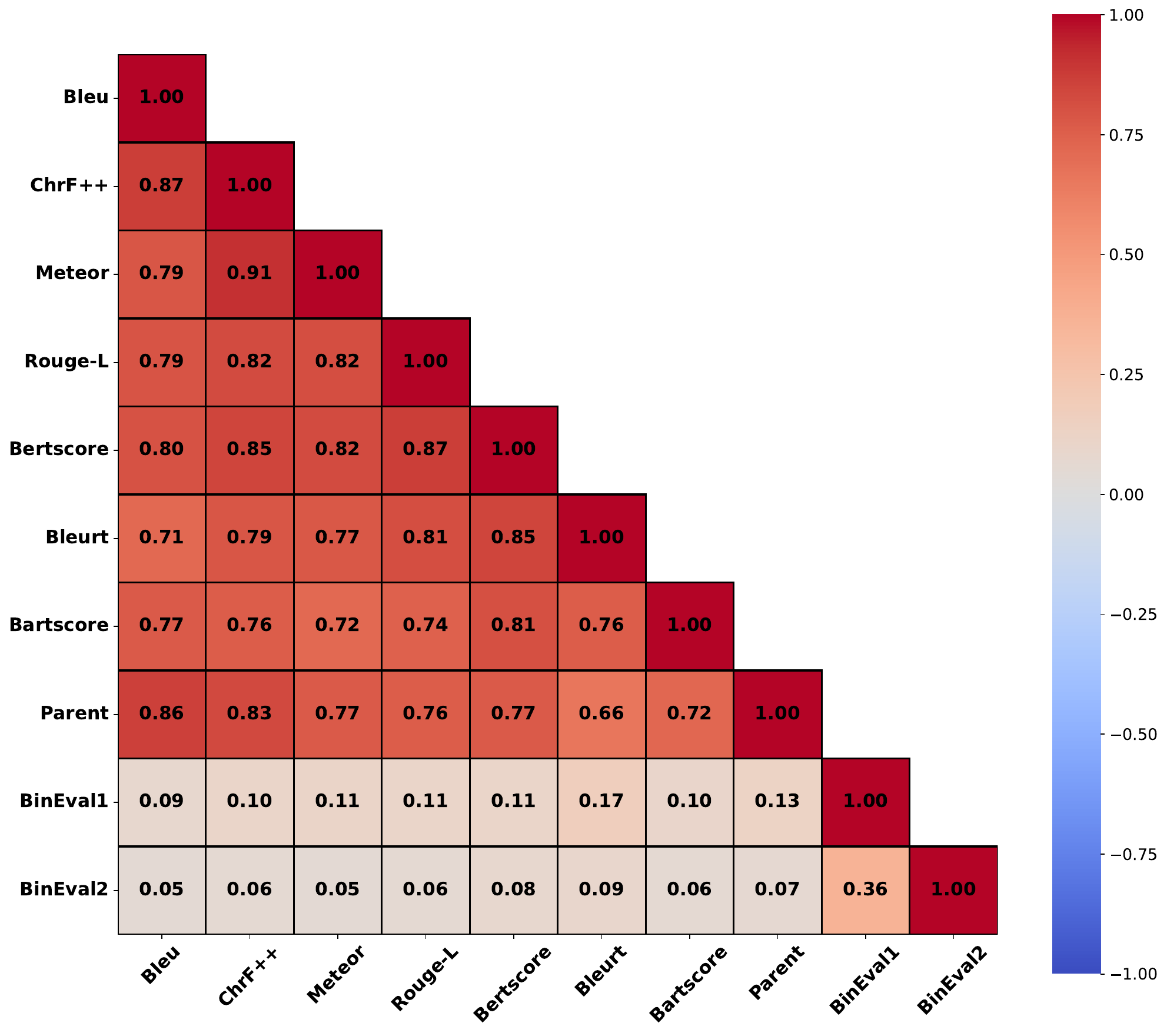}
        \caption{Nemotron-3-Ultra}
        \label{fig:nemo_correlation}
    \end{subfigure}
    \hfill
    \begin{subfigure}[t]{0.48\textwidth}
        \centering
        \includegraphics[width=\linewidth]{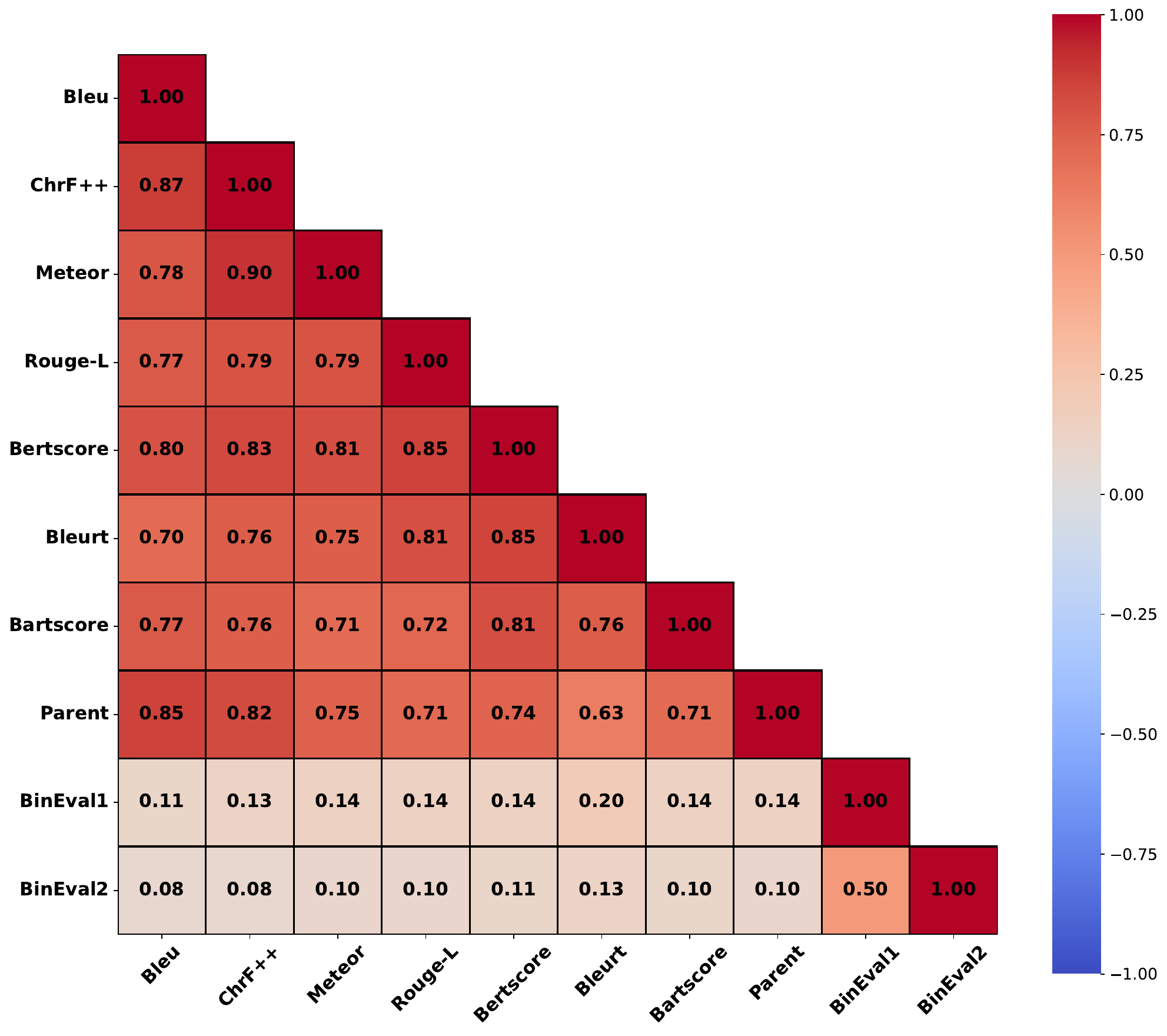}
        \caption{Granite-4.2-30B}
        \label{fig:granite_correlation}
    \end{subfigure}

    \vspace{0.5em}

    \begin{subfigure}[t]{0.48\textwidth}
        \centering
        \includegraphics[width=\linewidth]{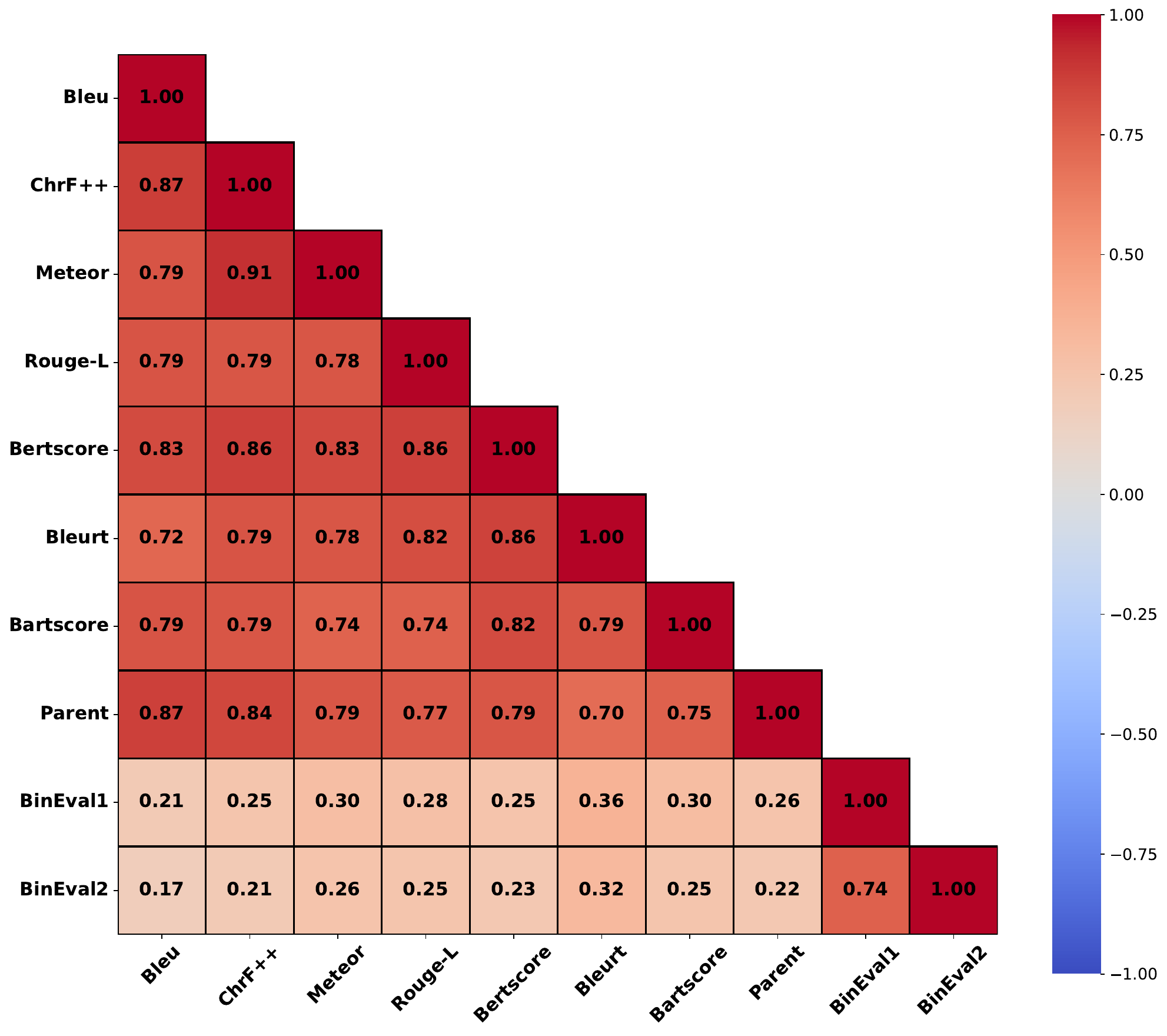}
        \caption{T5Gemma2 4B-4B}
        \label{fig:t5gemma_correlation}
    \end{subfigure}
    \hfill
    \begin{subfigure}[t]{0.48\textwidth}
        \centering
        \includegraphics[width=\linewidth]{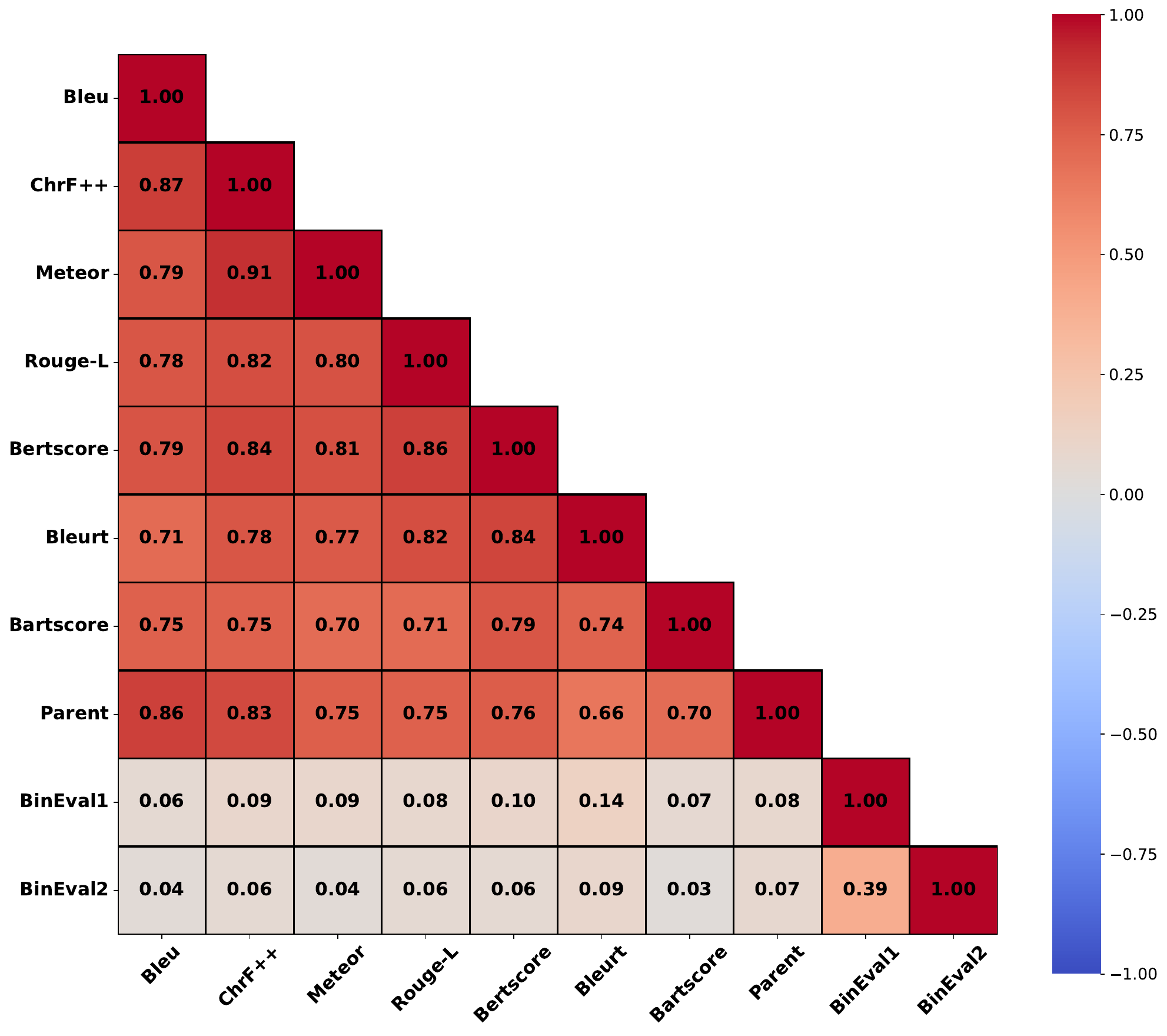}
        \caption{Gemma-4-31B-it}
        \label{fig:gemma_correlation}
    \end{subfigure}

    \caption{Spearman correlations between evaluation metrics for each model's generated outputs on WebNLG17. BinEval1 and BinEval2 denote faithfulness scores assigned by DeepSeek-V4.1-Flash and GLM-5.3-Flash, respectively.}
    \label{fig:model_correlation_heatmaps}
\end{figure*}

Since the evaluation metrics used in this work assess generated text from different perspectives, it is important to analyse them in detail. Figure \ref{fig:model_correlation_heatmaps} present the Spearman correlation between the evaluated metrics for all six model outputs on WebNLG17 dataset. Overall, the reference-based NLG metrics exhibit a strong correlation among themselves with correlation of 0.62 to 0.92, for all model outputs which drops to 0.36 to 0.06 with LLM judge: BinEval1 (Faithfulness using DeepSeek-V-4.1-Flash) with BLEURT correlating the most with it, mostly in case of fine-tuned models. The same goes for BinEval2 (Faithfulness using GLM-5.3-Flash). The two judge LLMs also correlate best with each other on fine-tuned models with correlation of 0.74 on T5-Gemma2 and of 0.67 on T5-large with least corelation value on GPT-5-mini. In addition to correlation between metrics,  Table \ref{tab:semantically similar} presents some examples where the surface-level metrics report a low score compared to semantic metrics because of the difference in textual framing, thus showing that the surface-level metrics, as commonly used alone for evaluation, were never sufficient to assess the correctness of the generated text.

\section{Conclusion}
In this paper, we show that the prompted decoder-only models stand close to, and in some cases (like on WebNLG20) even slightly ahead of, the fine-tuned baseline T5-large on reference-based evaluation, while clearly outperforming it on source-grounded evaluation; the LoRA-tuned encoder-decoder behaves like a fully fine-tuned T5 model on both. To better understand this, we examined how our LLM-based BinEval evaluations relate to the reference-based metrics: BinEval scores do not correlate strongly with the reference-based metric families, which themselves correlate well with one another across all model outputs.  We find that the judges reproduce human judgements closely on hard failure cases such as contradictions and entity corruption, but often fail to flag unsupported elaborations in the output text. Despite this latter limitation, the BinEval evaluations still support our conclusions that the prompted models used in this study contain fewer contradictions and triple omissions than the fine-tuned baseline. We also got evidence that the fine-tuned models become more prone to graph-grounded errors as the graph size increases compared to the prompted models. Current benchmarks have graphs of at most seven triples and texts of a few sentences; as the size and complexity of the input subgraphs increase, we expect both factual errors, omissions and ungrounded elaboration to become more prevalent, making their systematic detection essential, and no current automatic evaluation is able to reliably detect the latter. Thus, evaluating report-scale generation needs a larger aligned dataset and evaluation protocols that detect ungrounded elaboration.

\FloatBarrier
\section*{Limitations}
Our experiments results on T5 fine-tuning are not numerically identical to all the published values because of differences in evaluation protocol(Table \ref{tab:published comparison}). The details of our evaluation settings is provided in Section \ref{Evaluation}. Our experiments on decoder-only models rely on single generation runs per model, so we cannot rule out variance across repeated samples. Our fine-tuned pole is also limited to T5-large and a single LoRA-tuned model; extending fine-tuning to a decoder-only architecture is left to future work. For human evaluation, a single annotator rated outputs, with a second author adjudicating every disputed verdict with the LLM judges’ explanations visible; this visibility may have anchored the second author’s adjudication toward the judges’ reasoning rather than serving as a fully independent check. The annotation also covered only one prompted system (Nemotron-3-Ultra) and the Faithfulness and Coverage dimensions, leaving Fluency unannotated. With only 110 graphs annotated from a total of 1862 samples on WebNLG17, the resulting counts are small, and the reported agreement statistics carry wide confidence intervals. Because errors missed by both the annotator and the judges remain undetected, the judge recall figures we reported should be interpreted as upper bounds rather than exact values. We also want to highlight that the benchmark datasets WebNLG17, WebNLG20 and DART are not independent of each other by construction because DART incorporates WebNLG17 data directly, and WebNLG20 extends WebNLG17’s underlying triple set, so we cannot rule out the possibility of dataset memorization by judge LLMs. Finally, as we mentioned earlier, GPTScore using DeepSeek-V4.1-Flash has remained inconclusive; this study does not include an analysis of the reason for the same.

\bibliography{custom}

@inproceedings{gardent2017webnlg,
  title={The WebNLG challenge: Generating text from RDF data},
  author={Gardent, Claire and Shimorina, Anastasia and Narayan, Shashi and Perez-Beltrachini, Laura},
  booktitle={Proceedings of the 10th international conference on natural language generation},
  pages={124--133},
  year={2017}
}

@inproceedings{ferreira20202020,
  title={The 2020 bilingual, bi-directional WebNLG+ shared task: Overview and evaluation results (WebNLG+ 2020)},
  author={Ferreira, Thiago Castro and Gardent, Claire and Ilinykh, Nikolai and Van Der Lee, Chris and Mille, Simon and Moussallem, Diego and Shimorina, Anastasia},
  booktitle={Proceedings of the 3rd International Workshop on Natural Language Generation from the Semantic Web (WebNLG+)},
  pages={55--76},
  year={2020}
}

@inproceedings{nan2021dart,
  title={Dart: Open-domain structured data record to text generation},
  author={Nan, Linyong and Radev, Dragomir and Zhang, Rui and Rau, Amrit and Sivaprasad, Abhinand and Hsieh, Chiachun and Tang, Xiangru and Vyas, Aadit and Verma, Neha and Krishna, Pranav and others},
  booktitle={Proceedings of the 2021 Conference of the North American Chapter of the Association for Computational Linguistics: Human Language Technologies},
  pages={432--447},
  year={2021}
}

@article{li2024unifying,
  title={Unifying structured data as graph for data-to-text pre-training},
  author={Li, Shujie and Li, Liang and Geng, Ruiying and Yang, Min and Li, Binhua and Yuan, Guanghu and He, Wanwei and Yuan, Shao and Ma, Can and Huang, Fei and others},
  journal={Transactions of the Association for Computational Linguistics},
  volume={12},
  pages={210--228},
  year={2024},
  publisher={MIT Press One Broadway, 12th Floor, Cambridge, Massachusetts 02142, USA~…}
}

@inproceedings{yuan2023evaluating,
  title={Evaluating generative models for graph-to-text generation},
  author={Yuan, Shuzhou and F{\"a}rber, Michael},
  booktitle={Proceedings of the 14th International Conference on Recent Advances in Natural Language Processing},
  pages={1256--1264},
  year={2023}
}

@inproceedings{he2025evaluating,
  title={Evaluating and improving graph to text generation with large language models},
  author={He, Jie and Yang, Yijun and Long, Wanqiu and Xiong, Deyi and Guti{\'e}rrez-Basulto, V{\'\i}ctor and Pan, Jeff Z},
  booktitle={Proceedings of the 2025 Conference of the Nations of the Americas Chapter of the Association for Computational Linguistics: Human Language Technologies (Volume 1: Long Papers)},
  pages={10219--10244},
  year={2025}
}

@inproceedings{ribeiro2021investigating,
  title={Investigating pretrained language models for graph-to-text generation},
  author={Ribeiro, Leonardo FR and Schmitt, Martin and Sch{\"u}tze, Hinrich and Gurevych, Iryna},
  booktitle={Proceedings of the 3rd Workshop on Natural Language Processing for Conversational AI},
  pages={211--227},
  year={2021}
}

@inproceedings{papineni2002bleu,
  title={Bleu: a method for automatic evaluation of machine translation},
  author={Papineni, Kishore and Roukos, Salim and Ward, Todd and Zhu, Wei-Jing},
  booktitle={Proceedings of the 40th annual meeting of the Association for Computational Linguistics},
  pages={311--318},
  year={2002}
}

@inproceedings{banerjee2005meteor,
  title={METEOR: An automatic metric for MT evaluation with improved correlation with human judgments},
  author={Banerjee, Satanjeev and Lavie, Alon},
  booktitle={Proceedings of the acl workshop on intrinsic and extrinsic evaluation measures for machine translation and/or summarization},
  pages={65--72},
  year={2005}
}

@inproceedings{popovic2017chrf++,
  title={chrF++: words helping character n-grams},
  author={Popovi{\'c}, Maja},
  booktitle={Proceedings of the second conference on machine translation},
  pages={612--618},
  year={2017}
}

@article{yuan2021bartscore,
  title={Bartscore: Evaluating generated text as text generation},
  author={Yuan, Weizhe and Neubig, Graham and Liu, Pengfei},
  journal={Advances in neural information processing systems},
  volume={34},
  pages={27263--27277},
  year={2021}
}

@inproceedings{lin2004rouge,
  title={Rouge: A package for automatic evaluation of summaries},
  author={Lin, Chin-Yew},
  booktitle={Text summarization branches out},
  pages={74--81},
  year={2004}
}

@inproceedings{sellam2020bleurt,
  title={BLEURT: Learning robust metrics for text generation},
  author={Sellam, Thibault and Das, Dipanjan and Parikh, Ankur},
  booktitle={Proceedings of the 58th annual meeting of the association for computational linguistics},
  pages={7881--7892},
  year={2020}
}

@inproceedings{dhingra2019handling,
  title={Handling divergent reference texts when evaluating table-to-text generation},
  author={Dhingra, Bhuwan and Faruqui, Manaal and Parikh, Ankur and Chang, Ming-Wei and Das, Dipanjan and Cohen, William},
  booktitle={Proceedings of the 57th annual meeting of the association for computational linguistics},
  pages={4884--4895},
  year={2019}
}

@article{zhang2019bertscore,
  title={Bertscore: Evaluating text generation with bert},
  author={Zhang, Tianyi and Kishore, Varsha and Wu, Felix and Weinberger, Kilian Q and Artzi, Yoav},
  journal={arXiv preprint arXiv:1904.09675},
  year={2019}
}

@article{singh2025openai,
  title={Openai gpt-5 system card},
  author={Singh, Aaditya and Fry, Adam and Perelman, Adam and Tart, Adam and Ganesh, Adi and El-Kishky, Ahmed and McLaughlin, Aidan and Low, Aiden and Ostrow, AJ and Ananthram, Akhila and others},
  journal={arXiv preprint arXiv:2601.03267},
  year={2025}
}

@inproceedings{song2018graph,
  title={A graph-to-sequence model for AMR-to-text generation},
  author={Song, Linfeng and Zhang, Yue and Wang, Zhiguo and Gildea, Daniel},
  booktitle={Proceedings of the 56th Annual Meeting of the Association for Computational Linguistics (Volume 1: Long Papers)},
  pages={1616--1626},
  year={2018}
}

@inproceedings{zhao2020bridging,
  title={Bridging the structural gap between encoding and decoding for data-to-text generation},
  author={Zhao, Chao and Walker, Marilyn and Chaturvedi, Snigdha},
  booktitle={Proceedings of the 58th annual meeting of the association for computational linguistics},
  pages={2481--2491},
  year={2020}
}

@inproceedings{guo2020cyclegt,
  title={Cyclegt: Unsupervised graph-to-text and text-to-graph generation via cycle training},
  author={Guo, Qipeng and Jin, Zhijing and Qiu, Xipeng and Zhang, Weinan and Wipf, David and Zhang, Zheng},
  booktitle={Proceedings of the 3rd international workshop on natural language generation from the semantic web (WebNLG+)},
  pages={77--88},
  year={2020}
}

@inproceedings{li2021few,
  title={Few-shot knowledge graph-to-text generation with pretrained language models},
  author={Li, Junyi and Tang, Tianyi and Zhao, Wayne Xin and Wei, Zhicheng and Yuan, Nicholas Jing and Wen, Ji-Rong},
  booktitle={Findings of the Association for Computational Linguistics: ACL-IJCNLP 2021},
  pages={1558--1568},
  year={2021}
}

@inproceedings{koncel2019text,
  title={Text generation from knowledge graphs with graph transformers},
  author={Koncel-Kedziorski, Rik and Bekal, Dhanush and Luan, Yi and Lapata, Mirella and Hajishirzi, Hannaneh},
  booktitle={Proceedings of the 2019 Conference of the North American Chapter of the Association for Computational Linguistics: Human Language Technologies, Volume 1 (Long and Short Papers)},
  pages={2284--2293},
  year={2019}
}

@inproceedings{duvsek2018findings,
  title={Findings of the E2E NLG challenge},
  author={Du{\v{s}}ek, Ond{\v{r}}ej and Novikova, Jekaterina and Rieser, Verena},
  booktitle={Proceedings of the 11th International Conference on Natural Language Generation},
  pages={322--328},
  year={2018}
}

@article{zhong2017seq2sql,
  title={Seq2sql: Generating structured queries from natural language using reinforcement learning},
  author={Zhong, Victor and Xiong, Caiming and Socher, Richard},
  journal={arXiv preprint arXiv:1709.00103},
  year={2017}
}

@inproceedings{pasupat2015compositional,
  title={Compositional semantic parsing on semi-structured tables},
  author={Pasupat, Panupong and Liang, Percy},
  booktitle={Proceedings of the 53rd Annual Meeting of the Association for Computational Linguistics and the 7th International Joint Conference on Natural Language Processing (Volume 1: Long Papers)},
  pages={1470--1480},
  year={2015}
}

@inproceedings{shi2023hallucination,
  title={Hallucination mitigation in natural language generation from large-scale open-domain knowledge graphs},
  author={Shi, Xiao and Zhu, Zhengyuan and Zhang, Zeyu and Li, Chengkai},
  booktitle={Proceedings of the 2023 Conference on Empirical Methods in Natural Language Processing},
  pages={12506--12521},
  year={2023}
}

@article{wang2022text,
  title={Text embeddings by weakly-supervised contrastive pre-training},
  author={Wang, Liang and Yang, Nan and Huang, Xiaolong and Jiao, Binxing and Yang, Linjun and Jiang, Daxin and Majumder, Rangan and Wei, Furu},
  journal={arXiv preprint arXiv:2212.03533},
  year={2022}
}

@inproceedings{carbonell1998use,
  title={The use of MMR, diversity-based reranking for reordering documents and producing summaries},
  author={Carbonell, Jaime and Goldstein, Jade},
  booktitle={Proceedings of the 21st annual international ACM SIGIR conference on Research and development in information retrieval},
  pages={335--336},
  year={1998}
}

@inproceedings{post2018call,
  title={A call for clarity in reporting BLEU scores},
  author={Post, Matt},
  booktitle={Proceedings of the third conference on machine translation: Research papers},
  pages={186--191},
  year={2018}
}

@inproceedings{iarosh2025reducing,
  title={On reducing factual hallucinations in graph-to-text generation using large language models},
  author={Iarosh, Dmitrii and Panchenko, Alexander and Salnikov, Mikhail},
  booktitle={Proceedings of the Workshop on Generative AI and Knowledge Graphs (GenAIK)},
  pages={43--53},
  year={2025}
}

@article{team2026gemma,
  title={Gemma 4 technical report},
  author={Team, Gemma and Abd, Sherif El and Aggarwal, Vaibhav and Algayres, Robin and Andreev, Alek and Bachem, Olivier and Ballantyne, Ian and Brick, Cormac and C{\u{a}}rbune, Victor and Casbon, Michelle and others},
  journal={arXiv preprint arXiv:2607.02770},
  year={2026}
}

@misc{granite2026,
  author       = {{IBM Research}},
  title        = {Granite 4.2 Language Models},
  year         = {2026},
  howpublished = {\url{https://huggingface.co/blog/ibm-granite/granite-4-2}},
  note         = {Accessed: 2026-07-25}
}

@article{zhang2025t5gemma,
  title={T5gemma 2: Seeing, reading, and understanding longer},
  author={Zhang, Biao and Suganthan, Paul and Liu, Ga{\"e}l and Philippov, Ilya and Dua, Sahil and Hora, Ben and Black, Kat and Martins, Gus and Sanseviero, Omar and Pathak, Shreya and others},
  journal={arXiv preprint arXiv:2512.14856},
  year={2025}
}

@article{blakeman2025nvidia,
  title={Nvidia nemotron 3: Efficient and open intelligence},
  author={Blakeman, Aaron and Grattafiori, Aaron and Basant, Aarti and Gupta, Abhibha and Khattar, Abhinav and Renduchintala, Adi and Vavre, Aditya and Shukla, Akanksha and Bercovich, Akhiad and Ficek, Aleksander and others},
  journal={arXiv preprint arXiv:2512.20856},
  year={2025}
}

@article{cho2026ask,
  title={Ask, Don't Judge: Binary Questions for Interpretable LLM Evaluation and Self-Improvement},
  author={Cho, Sangwoo and Chawla, Kushal and Cai, Pengshan and Liu, Zefang and Zhu, Chenyang and Zhang, Shi-Xiong and Sahu, Sambit},
  journal={arXiv preprint arXiv:2606.27226},
  year={2026}
}

@inproceedings{fu2024gptscore,
  title={Gptscore: Evaluate as you desire},
  author={Fu, Jinlan and Ng, See Kiong and Jiang, Zhengbao and Liu, Pengfei},
  booktitle={Proceedings of the 2024 conference of the north american chapter of the association for computational linguistics: Human language technologies (volume 1: Long papers)},
  pages={6556--6576},
  year={2024}
}

@article{xu2026deepseek,
  title={Deepseek-v4. 1-flash: Pushing the limits of kv cache compression},
  author={Xu, Anyi and Li, B and Lin, Bangcai and Xue, Bing and Xian, BingCheng and Xu, Bingzheng and Wu, Bochao and Zhang, Bowei and Deng, Boyi and Yu, CC and others},
  journal={arXiv preprint arXiv:2609.19969},
  year={2026}
}

@misc{glm5team2026glm5vibecodingagentic,
      title={GLM-5: from Vibe Coding to Agentic Engineering},
      author={GLM-5-Team and : and Aohan Zeng and Xin Lv and Zhenyu Hou and Zhengxiao Du and Qinkai Zheng and Bin Chen and Da Yin and Chendi Ge and Chenghua Huang and Chengxing Xie and Chenzheng Zhu and Congfeng Yin and Cunxiang Wang and Gengzheng Pan and Hao Zeng and Haoke Zhang and Haoran Wang and Huilong Chen and Jiajie Zhang and Jian Jiao and Jiaqi Guo and Jingsen Wang and Jingzhao Du and Jinzhu Wu and Kedong Wang and Lei Li and Lin Fan and Lucen Zhong and Mingdao Liu and Mingming Zhao and Pengfan Du and Qian Dong and Rui Lu and Shuang-Li and Shulin Cao and Song Liu and Ting Jiang and Xiaodong Chen and Xiaohan Zhang and Xuancheng Huang and Xuezhen Dong and Yabo Xu and Yao Wei and Yifan An and Yilin Niu and Yitong Zhu and Yuanhao Wen and Yukuo Cen and Yushi Bai and Zhongpei Qiao and Zihan Wang and Zikang Wang and Zilin Zhu and Ziqiang Liu and Zixuan Li and Bojie Wang and Bosi Wen and Can Huang and Changpeng Cai and Chao Yu and Chen Li and Chengwei Hu and Chenhui Zhang and Dan Zhang and Daoyan Lin and Dayong Yang and Di Wang and Ding Ai and Erle Zhu and Fangzhou Yi and Feiyu Chen and Guohong Wen and Hailong Sun and Haisha Zhao and Haiyi Hu and Hanchen Zhang and Hanrui Liu and Hanyu Zhang and Hao Peng and Hao Tai and Haobo Zhang and He Liu and Hongwei Wang and Hongxi Yan and Hongyu Ge and Huan Liu and Huanpeng Chu and Jia'ni Zhao and Jiachen Wang and Jiajing Zhao and Jiamin Ren and Jiapeng Wang and Jiaxin Zhang and Jiayi Gui and Jiayue Zhao and Jijie Li and Jing An and Jing Li and Jingwei Yuan and Jinhua Du and Jinxin Liu and Junkai Zhi and Junwen Duan and Kaiyue Zhou and Kangjian Wei and Ke Wang and Keyun Luo and Laiqiang Zhang and Leigang Sha and Liang Xu and Lindong Wu and Lintao Ding and Lu Chen and Minghao Li and Nianyi Lin and Pan Ta and Qiang Zou and Rongjun Song and Ruiqi Yang and Shangqing Tu and Shangtong Yang and Shaoxiang Wu and Shengyan Zhang and Shijie Li and Shuang Li and Shuyi Fan and Wei Qin and Wei Tian and Weining Zhang and Wenbo Yu and Wenjie Liang and Xiang Kuang and Xiangmeng Cheng and Xiangyang Li and Xiaoquan Yan and Xiaowei Hu and Xiaoying Ling and Xing Fan and Xingye Xia and Xinyuan Zhang and Xinze Zhang and Xirui Pan and Xu Zou and Xunkai Zhang and Yadi Liu and Yandong Wu and Yanfu Li and Yidong Wang and Yifan Zhu and Yijun Tan and Yilin Zhou and Yiming Pan and Ying Zhang and Yinpei Su and Yipeng Geng and Yong Yan and Yonglin Tan and Yuean Bi and Yuhan Shen and Yuhao Yang and Yujiang Li and Yunan Liu and Yunqing Wang and Yuntao Li and Yurong Wu and Yutao Zhang and Yuxi Duan and Yuxuan Zhang and Zezhen Liu and Zhengtao Jiang and Zhenhe Yan and Zheyu Zhang and Zhixiang Wei and Zhuo Chen and Zhuoer Feng and Zijun Yao and Ziwei Chai and Ziyuan Wang and Zuzhou Zhang and Bin Xu and Minlie Huang and Hongning Wang and Juanzi Li and Yuxiao Dong and Jie Tang},
      year={2026},
      eprint={2602.15763},
      archivePrefix={arXiv},
      primaryClass={cs.LG},
      url={https://arxiv.org/abs/2602.15763},
}

@article{colas2021eventnarrative,
  title={Eventnarrative: A large-scale event-centric dataset for knowledge graph-to-text generation},
  author={Colas, Anthony and Sadeghian, Ali and Wang, Yue and Wang, Daisy Zhe},
  journal={arXiv preprint arXiv:2111.00276},
  year={2021}
}

@inproceedings{agarwal2021knowledge,
  title={Knowledge graph based synthetic corpus generation for knowledge-enhanced language model pre-training},
  author={Agarwal, Oshin and Ge, Heming and Shakeri, Siamak and Al-Rfou, Rami},
  booktitle={Proceedings of the 2021 Conference of the North American Chapter of the Association for Computational Linguistics: Human Language Technologies},
  pages={3554--3565},
  year={2021}
}

@inproceedings{jin2020genwiki,
  title={GenWiki: A dataset of 1.3 million content-sharing text and graphs for unsupervised graph-to-text generation},
  author={Jin, Zhijing and Guo, Qipeng and Qiu, Xipeng and Zhang, Zheng},
  booktitle={Proceedings of the 28th International Conference on Computational Linguistics},
  pages={2398--2409},
  year={2020}
}

@inproceedings{lee2025checkeval,
  title={Checkeval: A reliable llm-as-a-judge framework for evaluating text generation using checklists},
  author={Lee, Yukyung and Kim, Joonghoon and Kim, Jaehee and Cho, Hyowon and Kang, Jaewook and Kang, Pilsung and Kim, Najoung},
  booktitle={Proceedings of the 2025 Conference on Empirical Methods in Natural Language Processing},
  pages={15771--15798},
  year={2025}
}

@article{efimov2026xqdt,
  title={XQDT: eXplainable and Quantitative Data-Text Alignment Metric with Feedback Signals},
  author={Efimov-Zhang, Kun and Song, Yifei and Gardent, Claire},
  journal={arXiv preprint arXiv:2608.29948},
  year={2026}
}

@inproceedings{huidrom2025assessing,
  title={Assessing Semantic Consistency in Data-to-Text Generation: A Meta-Evaluation of Textual, Semantic and Model-Based Metrics},
  author={Huidrom, Rudali and Lorandi, Michela and Mille, Simon and Thomson, Craig and Belz, Anja},
  booktitle={Proceedings of the 18th International Natural Language Generation Conference},
  pages={98--107},
  year={2025}
}

@article{hu2021lora,
  title={Lora: Low-rank adaptation of large language models},
  author={Hu, Edward J and Shen, Yelong and Wallis, Phillip and Allen-Zhu, Zeyuan and Li, Yuanzhi and Wang, Shean and Wang, Lu and Chen, Weizhu},
  journal={arXiv preprint arXiv:2106.09685},
  year={2021}
}

@inproceedings{koehn2004statistical,
  title={Statistical significance tests for machine translation evaluation},
  author={Koehn, Philipp},
  booktitle={Proceedings of the 2004 conference on empirical methods in natural language processing},
  pages={388--395},
  year={2004}
}

@inproceedings{liu2023g,
  title={G-eval: NLG evaluation using gpt-4 with better human alignment},
  author={Liu, Yang and Iter, Dan and Xu, Yichong and Wang, Shuohang and Xu, Ruochen and Zhu, Chenguang},
  booktitle={Proceedings of the 2023 conference on empirical methods in natural language processing},
  pages={2511--2522},
  year={2023}
}

@inproceedings{kim2026ontology,
  title={Ontology-free general-domain knowledge graph-to-text generation dataset synthesis using large language model},
  author={Kim, Daehui and Kang, Deokhyung and Ryu, Sangwon and Lee, Gary Geunbae},
  booktitle={Proceedings of the First Workshop on Structured Understanding, Retrieval, and Generation in the LLM Era (SURGeLLM 2026)},
  pages={52--69},
  year={2026}
}

@article{shazeer2017outrageously,
  title={Outrageously large neural networks: The sparsely-gated mixture-of-experts layer},
  author={Shazeer, Noam and Mirhoseini, Azalia and Maziarz, Krzysztof and Davis, Andy and Le, Quoc and Hinton, Geoffrey and Dean, Jeff},
  journal={arXiv preprint arXiv:1701.06538},
  year={2017}
}

@article{dao2024transformers,
  title={Transformers are SSMs: Generalized models and efficient algorithms through structured state space duality},
  author={Dao, Tri and Gu, Albert},
  journal={arXiv preprint arXiv:2405.21060},
  year={2024}
}

\appendix

\section{Error Analysis of simpler prompt setting}
\label{appendix:error analysis}
GPT-5-mini consistently produces fluent text; however, the quality of the generated output is highly dependent on the prompt provided to the model. We initially employed the prompt Generate text from graph:\{\(graph\)\},'' following the approach of \citet{yuan2023evaluating} for the WebNLG dataset. During preliminary experiments, we observed several recurring issues. The model frequently failed to recognize that "<H>", "<R>" and "<T>" were structural markers and instead interpreted them as literal entity identifiers, referring to the head, relation, and tail entities as "Entity H", "Entity R" and "Entity T", respectively. Moreover, the model produced text that does not make sense semantically but is consistent with the order of entities present in the graph. In addition, the GPT-5 models occasionally produced explanatory comments or intermediate reasoning along with the generated output. These observations are consistent with the error analysis reported by \citet{yuan2023evaluating}. Examples of these errors are presented in Table~\ref{tab:hallucination}. To mitigate these issues, we incorporated explicit instructions addressing these common failure cases within the detailed prompt shown in Table~\ref{tab:[prompt]}. We observed that the revised prompt was effective in eliminating these recurring errors from the generated outputs.

\begin{table*}[t!]
\centering
\small
\begin{tabular}{p{2.5cm} p{8cm}}
\hline
\textbf{WebNLG } & Example \\

\hline

\textbf{Graph} &
<H> Abel Hernandez <R> club <T> Uruguay national football team
\\
\textbf{Target} & Abel Hernandez's club is the Uruguay National football team.\\
\textbf{GPT-5-mini} &
Abel Hernandez is a club for the Uruguay national football team. (Note: If you intended a different relationship phrasing, e.g., ``Abel Hernandez plays for the Uruguay national football team,'' that would also be correct.)
\\

\hline
\textbf{Graph} &
<H> 66063 1998 RO1 <R> epoch <T> 2013 11 04
\\
\textbf{Target} & 66063 1998 RO1 has an epoch date of 4 November 2013.\\
\textbf{GPT-5-mini} &
From the graph, the subject H has the following relation: In 1998 RO1 66063, epoch R, with T 2013-11-04.
However, the given data is a bit ambiguous as RDF. A clearer fluent rendering could be:
- In 1998, the resource H has the identifier 66063 and is related to RO1 via the property named epoch, with the value 2013-11-04.
- The triple: (H, epoch, 2013-11-04) includes the additional contextual numbers 66063, 1998, RO1, and T.
If you can provide the intended predicate names and the exact meaning of 66063, RO1, and the T value, I can produce a more natural fluent sentence. For now a best guess:
- The resource H, identified by 66063, originated in 1998 under RO1 and has an epoch timestamp of 2013-11-04.
)
\\
\hline
\end{tabular}

\caption{Sample text generated by GPT-5-mini with the prompt \textit{"Generate text from graph"}.}
\label{tab:hallucination}
\end{table*}

\begin{table}[t]
\centering
\small
\begin{tabular}{p{0.58\columnwidth}cc}
\hline

\textbf{Model} &
\textbf{BLEU} &
\textbf{ChrF++}

\\

\hline

\textbf{T5-Large on WebNLG17 \cite{ribeiro2021investigating}}
& 59.70
& 75.40

\\

\textbf{ Our reproduced results on WebNLG17}
& 58.57
& 72.54

\\
\hline

\textbf{T5-Large on DART \cite{li2024unifying}}
& 50.66
& n/r

\\
\textbf{Our reproduced results on DART}
& 47.87
& 65.11
\\
\hline

\end{tabular}

\caption{Comparison of the published results on WebNLG17 and DART on T5-Large and our reported results. It should be noted that we did not reproduce the evaluation protocol followed by the cited works; the BLEU and ChrF++ implementations used by us are mentioned in section \ref{Evaluation}.}
\label{tab:published comparison}
\end{table}

\begin{table*}[t!]
\centering
\small
\resizebox{\textwidth}{!}{
\begin{tabular}{lcccccccc}
\hline

\textbf{Model} &
\textbf{BLEU} &
\textbf{ChrF++} &
\textbf{ROUGE-L} &
\textbf{BLEURT} &
\textbf{BERTScore} &
\textbf{BARTScore} &
\textbf{PARENT F1}
\\

\hline

\textbf{GPT-3\cite{yuan2023evaluating}}
& 20.36
& 57.95
& 45.64
& 13.39
& n/r
& n/r
& n/r
\\

\textbf{ LLaMA2-7b-chat fine-tuned on PlanGTG\cite{he2025evaluating}}
&46.35
& 63.92
& n/r
& n/r
& n/r
& -1.91
& n/r
\\

\textbf{GPT-5 mini (3-shot prompted)}
& 51.94
& 70.70*
& 70.16*
& 81.38*
& 0.86
& -1.11
& 0.64
\\
\textbf{Gemma-4-31B (3-shot prompted)}
& 53.74
& 71.67*
& 71.73*
& 82.01*
& 0.87
& -1.03
& 0.66
\\

\textbf{Granite-4.2-30B (3-shot prompted)}
& 52.86
& 71.48*
& 70.23*
& 81.26*
& 0.86
& -1.08
& 0.66
\\

\textbf{Nemotron-3-Ultra (3-shot prompted)}
& 55.51
& 71.69*
& 72.31*
& 82.09*
& 0.87
& -1.01
& 0.66
\\

\textbf{T5-Gemma2-4B-4B (LoRA tuned)}
& 55.42
& 71.35*
& 70.49*
& 81.10*
& 0.87
& -1.09
& 0.65
\\

\hline

\end{tabular}
}
\caption{Comparison of the recent LLMs with results reported in \cite{yuan2023evaluating} and \cite{he2025evaluating} on WebNLG17. *Scores are multiplied by 100 to match the scale of reported scores.}
\label{tab:comparison}
\end{table*}

\begin{table*}[t]
\centering
\small
\resizebox{\textwidth}{!}{
\begin{tabular}{lcccccccc}
You are a system that converts RDF triples from a knowledge graph into natural English text.\\

Instructions:\\
- The triples describe facts about entities.\\
- Each RDF triple is in the format: <H> subject <R> predicate <T> object where the tokens <H>, <R>, and <T> are structural markers and must NOT appear in the output.\\
- Do not refer to entities as "entity H", "entity R", or "entity T".
- Relations are undirected; express them in the most natural way.
- Combine all triples into a coherent paragraph.\\
- Use one or more fluent sentences if needed.\\
- Do NOT use bullet points, numbering, or lists.\\
- Do NOT add explanations or commentary.\\
- Output only the final text.\\

RDF triples:\\
source line
\end{tabular}
}

\caption{Prompt provided to all decoder-only models.}
\label{tab:[prompt]}

\end{table*}

\begin{table}[t]
\centering
\scriptsize
\setlength{\tabcolsep}{2.5pt}
\begin{tabular}{lcccc}
\toprule
& \multicolumn{4}{c}{Fluency} \\
\cmidrule(lr){2-5}
System
& \multicolumn{2}{c}{WebNLG17}
& \multicolumn{2}{c}{WebNLG20} \\
\cmidrule(lr){2-3}
\cmidrule(lr){4-5}
& DeepSeek & GLM & DeepSeek & GLM \\
\midrule
T5-Large  & 0.943 & 0.899 & 0.951 & 0.925\\
T5-Gemma2 & 0.938 & 0.895 & 0.949 & 0.926\\
GPT-5-mini & 0.961 & 0.934 & 0.966 & 0.958\\
Granite-4.2 & 0.949 & 0.915 & 0.945 & 0.942\\
Nemotron-3-Ultra & 0.972 & 0.953 & 0.975 & 0.969\\
Gemma-4 & 0.970 & 0.949 & 0.979 & 0.974\\
\bottomrule
\end{tabular}
\caption{Fluency scores by system, dataset, and LLM judge. DeepSeek stands for DeepSeek-V-4.1-Flash and GLM stands for GLM-5.3-Flash.}
\label{tab:fluency}
\end{table}

\section{Bineval and GPTScore details}
\label{appendix: Bineval and gptscore details}
In this section, we provide the details of BinEval score and GPTScore evaluation. BinEval Score has been used to judge the outputs of six models on three datasets as mentioned earlier, for semantic faithfulness of the generated text with the graph, graph triple coverage in text and fluency of the text. For evaluating semantic faithfulness, we used the following system prompt:\\
\textit{Evaluation input format: The input graph is a knowledge graph represented as a sequence of one or more linearized triples.Each triple has the following structure: <H> Head <R> Relation <T> Tail where "<H>", "<R>", and "<T>" are literal strings that indicate the head, relation, and tail of the triple, respectively. The input graph is followed by a candidate output text that is generated from the graph. Your task is to evaluate whether the output text accurately reflects the information in the input graph by answering the provided questions. For each question, you should provide a JSON object with two fields: "satisfied" and "explanation". The "satisfied" field should be a boolean value indicating whether the
output text satisfies the question, and the "explanation" field should be a string providing a brief explanation of your answer. Judge the text only according to the respective question."}.\\
For the dataset Dart, we made some changes in the prompt to help the evaluator LLM understand the Dart dataset graph format better:\\
\textit{Evaluation input format: The input graph is a knowledge graph represented as a sequence of one or more linearized triples.Each triple has the following structure: <H> Head <R> Relation <T> Tail where "<H>", "<R>", and "<T>" are literal strings that indicate the head, relation, and tail of the triple, respectively. Some graphs are derived from tables and may contain special bracketed markers instead of real-world entities. "[TABLECONTEXT]" represents the data table or record as a whole - it is not a specific entity and should never be treated as one. When the relation is "[TITLE]" or "[title]", the triple's tail is the title of the table or record, not a literal string that must appear verbatim (e.g. <H> [TABLECONTEXT] <R> [title] <T> A Severed Wasp; means the table's title/subject is "A Severed Wasp" - this is content the candidate text may describe). The input graph is followed by a candidate output text that is generated from the graph. Your task is to evaluate whether the output text accurately reflects the information in the input graph by answering the provided questions. For each question, you should provide a JSON object with two fields: "satisfied" and "explanation". The "satisfied" field should be a boolean value indicating whether the output text satisfies the question, and the "explanation" field should be a string providing a brief explanation of your answer. Judge the text only according to the respective question.}
The prompt is accompanied with three questions and corresponding violation examples presented in table\ref{tab:sf}. The violation examples in the prompts are not taken from the test sets.\\
For the evaluation of fluency of the outputs, we used the following prompt to define the task:\\
\textit{Evaluation input format: You will be provided a candidate output text. Your task is to evaluate whether the output text is in fluent English by answering the provided question. For each question, you should provide a JSON object with two fields: "satisfied" and "explanation". The "satisfied" field should be a boolean value indicating whether the output text satisfies the question, and the "explanation" field should be a string providing a short explanation of your answer. Judge the text only according to the respective question.}\\
The three questions and corresponding violation examples are presented in table\ref{tab:fl}.\\
For Coverage, we converted the input graph into an ordered list of triples as input to the evaluator and for each triple, asked the evaluator to answer the question "\textit{Does the text describe the information present in this triple?}" and recorded the binary answer for each triple and obtained the ratio of sum of the scores to the total triples.
For GPTScore, we evaluate the likelihood scores for the generated text given the input graph and instruction prompt describing the task: \textit{"Generate a natural-language text description of the following knowledge graph"}.
\begin{table*}[t!]
\centering
\small
\begin{tabular}{p{2.5cm} p{8cm}}
\hline
\textbf{Question 1 } & Is every fact stated in the text supported by the input graph? \\
\textbf{Violation Example } & Input graph: <H> A Loyal Character Dancer <R> country <T> United States <H> United States <R> ethnic Group <T> Asian Americans \\
\textbf{} & Candidate text: A Loyal Character Dancer is published in English in the United States where the Asian Americans are one of the ethnic groups. \\
\textbf{} & Explanation: The text states that A Loyal Character Dancer is published in English, which is not supported by the input graph.  \\
\hline
\textbf{Question 2} & Are all entity names, attribute values, and numerical values reproduced correctly from the graph? \\
\textbf{Violation Example } & Input graph: <H> A Loyal Character Dancer <R> country <T> United States <H> United States <R> ethnic Group <T> Asian Americans \\
\textbf{} & Candidate text: The Loyal Character Dancers are from the United States where the Asian Americans are one of the ethnic groups. \\
\textbf{} & Explanation: The text mentions 'The Loyal Character Dancers' instead of 'A Loyal Character Dancer', which is not supported by the input graph.  \\
\hline
\textbf{Question 3 } & Are all relations between entities expressed correctly without reversal or contradiction? \\
\textbf{Violation Example } & Input graph: Input graph: <H> A Severed Wasp <R> author <T> Madeleine L'Engle \\
\textbf{} & Candidate text: Candidate text: A Severed Wasp is the author of Madeleine L'Engle. \\
\textbf{} & Explanation: The text reverses the relation, stating that A Severed Wasp is the author of Madeleine L'Engle, which contradicts the input graph.  \\

\hline
\end{tabular}
\caption{Questions and violation example provided to the judge LLMs for evaluating Faithfulness using BinEval framework}
\label{tab:sf}

\end{table*}

\begin{table*}[t!]
\centering
\small
\begin{tabular}{p{2.5cm} p{8cm}}
\hline
\textbf{Question 1 } & Is the text easy to read and understand? \\
\textbf{Violation Example } & Candidate text: Mount Early is near the head of Scott Glacier on the west side. It is high and stands north of D'Angelo Bluff. It is a solitary volcanic cone. \\
\textbf{} & Explanation: The text presents the information in a complex manner introducing the location first, then the height and type of landform in the end, which makes it hard to follow. \\
\hline
\textbf{Question 2} & Does the text avoid awkward or repetitive wording? \\
\textbf{Violation Example } & Candidate text: Interior is an 1895 play. It is a play in rhymed dialogue written by Belgian playwright Maurice Maeterlinck. \\
\textbf{} & Explanation: The text contains unnecessary repetition of the fact that Interior is a play. \\

\hline
\textbf{Question 3 } & Are the sentences in the text connected in a coherent manner and progress naturally? \\
\textbf{Violation Example } & Candidate text: The 1966 Arab Cup Final was a football match. The Al-Kashafa Stadium is in Baghdad, Iraq. The match took place on 10 April 1966, to determine the winner of the 1966 Arab Cup. It took place at the Al-Kashafa Stadium. \\
\textbf{} & Explanation: The text first defines the match and talks about the stadium in the next sentence while making the relation between the match and the stadium clear much later. \\
\hline
\end{tabular}
\caption{Questions and violation example provided to the judge LLMs for evaluating Fluency using BinEval framework}
\label{tab:fl}

\end{table*}

\begin{table*}[t!]
\centering
\small
\begin{tabular}{p{2.5cm} p{8cm}}
\hline
\textbf{WebNLG } & Example \\

\hline

\textbf{Graph} &
<H> Austria <R> language <T> Austrian German
\\
\textbf{Target 1} & In Austria, the language is Austrian German.\\
\textbf{Target 2} & The language in Austria is Austrian German.\\
\textbf{Target 3} & The spoken language of Austria is Austrian German.\\
\textbf{GPT-5-mini} & Austrian German is a language spoken in Austria.
\\

\hline
\textbf{Graph} & <H> Binignit <R> ingredient <T> Sweet potato <H> Sweet potato <R> division <T> Flowering plant <H> Binignit <R> main Ingredients <T> Sago <H> Sweet potato <R> order <T> Solanales
\\
\textbf{Target 1} & The binignit recipe calls for sweet potatoes and sago. The sweet potato belongs to the flowering plant and falls under the order of Solanales.\\
\textbf{Target 2} & Sago and Sweet potato are ingredients in Binignit. Sweet potato belongs to the Solanales order of flowering plants.
\\
\textbf{GPT-5-mini} &
Binignit includes sago as a main ingredient and sweet potato as an ingredient; the sweet potato is a flowering plant in the order Solanales.
\\
\hline
\end{tabular}

\caption{Some examples where the text generated by GPT-5-mini conveys the same meaning as the graph and references, but the surface overlap with all the references is low.}
\label{tab:semantically similar}

\end{table*}

\begin{table*}[t!]
\centering
\small
\begin{tabular}{p{2.5cm} p{8cm}}
\hline
\textbf{WebNLG } & Example \\

\hline

\textbf{Graph} &
<H> Above the Veil <R> media Type <T> Print
\\

\textbf{T5-large output} & Above the Veil was produced in print
\\
\textbf{GPTScore} & -5.500\\
\hline
\textbf{Graph} & <H> Bananaman <R> broadcasted By <T> BBC
\\

\textbf{T5-large output} & Bananaman was broadcast by the BBC
\\
\textbf{GPTScore} & -3.699 \\
\hline
\textbf{Graph} & <H> Asam pedas <R> country <T> Malaysia <H> Malaysia <R> ethnic Group <T> Malaysian Malay <H> Sumatra <R> ethnic Group <T> Batak <H> Asam pedas <R> region <T> Sumatram.
\\

\textbf{T5-large output} & Asam pedas is a dish from Sumatra and Malaysia. The Batak are an ethnic group from Sumatra. \\
\textbf{GPTScore} & -1.5130 \\
\hline
\end{tabular}

\caption{Some examples where the GPTScore value is low despite the output being correct and high despite the output misses one triple}.
\label{tab:gptscore}

\end{table*}

\begin{table*}[t]
\centering
\scriptsize
\setlength{\tabcolsep}{3pt}
\renewcommand{\arraystretch}{1.15}
\begin{tabular}{ll*{12}{c}}
\toprule
\textbf{Triples} & \textbf{$n$}
& \multicolumn{2}{c}{\textbf{T5-Large}}
& \multicolumn{2}{c}{\textbf{T5-Gemma2 (LoRA)}}
& \multicolumn{2}{c}{\textbf{Gemma-4-31B}}
& \multicolumn{2}{c}{\textbf{Granite-4.2-30B}}
& \multicolumn{2}{c}{\textbf{GPT-5 mini}}
& \multicolumn{2}{c}{\textbf{Nemotron-3-Ultra}} \\
\cmidrule(lr){3-4}
\cmidrule(lr){5-6}
\cmidrule(lr){7-8}
\cmidrule(lr){9-10}
\cmidrule(lr){11-12}
\cmidrule(lr){13-14}
& &
\textbf{DeepSeek} & \textbf{GLM}
& \textbf{DeepSeek} & \textbf{GLM}
& \textbf{DeepSeek} & \textbf{GLM}
& \textbf{DeepSeek} & \textbf{GLM}
& \textbf{DeepSeek} & \textbf{GLM}
& \textbf{DeepSeek} & \textbf{GLM} \\
\midrule

1& 454 & 4.6 & 3.3 & 4.6 & 3.5 & 2.0 & 0.9 & 3.7 & 1.1 & 3.7 & 1.5 & 2.6 & 1.3 \\

2& 349 & 12.6 & 7.7 & 14.6 & 8.3 & 4.9 & 3.2 & 8.0 & 5.4 & 4.3 & 1.5 & 8.0 & 3.2\\

3& 386 & 21.5 & 15.0 & 27.2 & 17.6 & 6.0 & 2.1 & 12.4 & 8.3 & 10.1 & 2.8 & 10.6 & 4.1 \\

4& 363 & 23.7 & 15.4 & 34.2 & 24.5 & 5.5 & 2.8 & 12.1 & 4.4 & 11.8 & 4.7 & 11.3 & 4.7 \\

5& 265 & 30.9 & 24.2 & 40.0 & 32.1 & 5.3 & 4.2 & 13.2 & 7.2 & 11.3 & 3.4 & 12.5 & 5.7 \\

6 or 7& 45 & 22.2 & 11.1 & 55.6 & 42.2 & 2.2 & 2.2 & 17.8 & 17.8 & 20.0 & 8.9 & 8.9 & 2.2 \\
\bottomrule
\end{tabular}
\caption{Faithfulness by number of input triples on WebNLG17, evaluated by DeepSeek-V-4.1-Flash and GLM-5.3-Flash. \(n\) denotes number of graphs}
\label{tab:faithfulness_by_triples}
\end{table*}

\begin{table*}[t]
\centering
\scriptsize
\setlength{\tabcolsep}{3pt}
\renewcommand{\arraystretch}{1.1}
\begin{tabular}{cc*{12}{c}}
\toprule
\textbf{Triples} & \textbf{$n$}
& \multicolumn{2}{c}{\textbf{T5-Large}}
& \multicolumn{2}{c}{\textbf{T5-Gemma2 (LoRA)}}
& \multicolumn{2}{c}{\textbf{Gemma-4-31B}}
& \multicolumn{2}{c}{\textbf{Granite-4.2-30B}}
& \multicolumn{2}{c}{\textbf{GPT-5 mini}}
& \multicolumn{2}{c}{\textbf{Nemotron-3-Ultra}} \\
\cmidrule(lr){3-4}
\cmidrule(lr){5-6}
\cmidrule(lr){7-8}
\cmidrule(lr){9-10}
\cmidrule(lr){11-12}
\cmidrule(lr){13-14}
& &
\textbf{DeepSeek} & \textbf{GLM}
& \textbf{DeepSeek} & \textbf{GLM}
& \textbf{DeepSeek} & \textbf{GLM}
& \textbf{DeepSeek} & \textbf{GLM}
& \textbf{DeepSeek} & \textbf{GLM}
& \textbf{DeepSeek} & \textbf{GLM} \\
\midrule

1& 454 & 0.9 & 0.4 & 0.7 & 0.9 & 0.0 & 0.0 & 0.4 & 0.4 & 0.2 & 0.0 & 0.0 & 0.0 \\

2& 349 & 5.4 & 3.4 & 4.9 & 4.0 & 2.0 & 1.4 & 2.3 & 1.1 & 2.0 & 0.6 & 1.4 & 1.1\\

3& 386 & 13.2 & 12.4 & 14.0 & 11.1 & 2.8 & 1.8 & 5.2 & 3.9 & 2.6 & 1.0 & 4.1 & 1.3 \\

4& 363 & 22.3 & 17.1 & 20.1 & 14.3 & 3.6 & 1.4 & 4.4 & 2.8 & 3.3 & 2.2 & 4.7 & 3.6 \\

5& 265 & 34.3 & 29.8 & 35.5 & 29.8& 6.4 & 2.3 & 6.0 & 2.6 & 5.3 & 1.5 & 6.8 & 1.1 \\

6 or 7& 45 & 28.9 & 28.9 & 40.0 & 31.1 & 2.2 & 0.0 & 8.9 & 4.4 & 2.2 & 0.0 & 6.7 & 0.0 \\
\bottomrule
\end{tabular}
\caption{Coverage scores by number of input triples on WebNLG17 dataset by LLM judges. \(n\) denotes number of graphs}
\label{tab:coverage_by_triples}
\end{table*}

\end{document}